\documentclass{article} 
\usepackage[preprint]{conference}

\usepackage{microtype}
\usepackage{hyperref}
\usepackage{url}
\usepackage{booktabs}
\usepackage{wrapfig}
\usepackage{lineno}
 \usepackage[utf8]{inputenc}
 \usepackage[T1]{fontenc}

\usepackage{wrapfig}
\usepackage{booktabs} 

\newcommand{\coloredmatrix}[4]{
\begingroup
\setlength{\tabcolsep}{9.5pt} 
\renewcommand{\arraystretch}{1} 
    \begin{tabular}{cc}
        \cellcolor{red!30}{#1} & \cellcolor{blue!30}{#2} \\
        \cellcolor{green!30}{#3} & \cellcolor{yellow!30}{#4}
    \end{tabular}
\endgroup
}

\newcommand{\coloredmatrixdelta}[8]{
\begingroup
\setlength{\tabcolsep}{0pt} 
\renewcommand{\arraystretch}{1} 
\begin{tabular}{cc}
 \cellcolor{red!30}{#1}{#2} &
 \cellcolor{blue!30}{#3}{#4} \\
 \cellcolor{green!30}{#5}{#6} &
 \cellcolor{yellow!30}{#7}{#8}
\end{tabular}
\endgroup
}
\newcommand{\up}[1]{
{$_{\uparrow #1}$}
}
\newcommand{\down}[1]{
{$_{\downarrow #1}$}
}
\newcommand{\same}{
{\hspace{0.1in}$_=$}
}

\newcommand{\notapplicable}{\hspace{0.3in}N/A}
\newcommand{\inferenceconditions}{inference methods}

\usepackage{tabularx}
\usepackage{amsmath}
\usepackage{graphicx}
\usepackage{subcaption}
\usepackage{enumitem}
\usepackage{tcolorbox}
\usepackage{colortbl}
\usepackage{color}
\usepackage{diagbox}

\definecolor{darkblue}{rgb}{0, 0, 0.5}
\hypersetup{colorlinks=true, citecolor=darkblue, linkcolor=darkblue, urlcolor=darkblue}

\title{How well can LLMs provide planning feedback in grounded environments?}

\author{Yuxuan Li \\
David R. Cheriton School of Computer Science \\
University of Waterloo\\
Waterloo, ON, N2L 3G1, Canada \\
\texttt{yxuuan.li1@uwaterloo.ca} \\
\And
Victor Zhong \\
David R. Cheriton School of Computer Science \\
University of Waterloo\\
Waterloo, ON, N2L 3G1, Canada \\
\texttt{victor.zhong@uwaterloo.ca} \\
}

\newcommand{\feedbackmodel}{{F}}
\newcommand{\ybinary}{{y_{\rm bin}}}
\newcommand{\yaction}{{y_{\rm act}}}
\newcommand{\ypref}{{y_{\rm pref}}}
\newcommand{\ygoal}{{y_{\rm goal}}}
\newcommand{\ydelta}{{y_\Delta}}

\begin{document}

\ifanonymityplaceholdersubmission
\linenumbers
\fi

\maketitle

\begin{abstract}
Learning to plan in grounded environments typically requires carefully designed reward functions or high-quality annotated demonstrations.
Recent works show that pretrained foundation models, such as large language models (LLMs) and vision language models (VLMs), capture background knowledge helpful for planning, which reduces the amount of reward design and demonstrations needed for policy learning.
We evaluate how well LLMs and VLMs provide feedback across symbolic, language, and continuous control environments.
We consider prominent types of feedback for planning including binary feedback, preference feedback, action advising, goal advising, and delta action feedback.
We also consider \inferenceconditions~that impact feedback performance, including in-context learning, chain-of-thought, and access to environment dynamics.
We find that foundation models can provide diverse high-quality feedback across domains.
Moreover, larger and reasoning models consistently provide more accurate feedback, exhibit less bias, and benefit more from enhanced \inferenceconditions.
Finally, feedback quality degrades for environments with complex dynamics or continuous state spaces and action spaces.
\end{abstract}

\section{Introduction}

Training instruction following agents in grounded environments often requires a large number of trials~\citep{mnih2013dqn,schulman2017ppo} or a large set of high-quality annotated demonstrations~\citep{Stepputtis2020-language-conditionedtasks,robomimic2021,saycan2022arxiv}.
Recent progress in large-scale pretraining shows that large language models (LLMs) capture a significant amount of world-knowledge that are helpful for reasoning and planning~\citep{brown2020language,saycan2022arxiv}.
While these models struggle as direct policies in grounded environments~\citep{robomimic2021,xie2024osworld}, they can provide helpful judgment and feedback on agent behaviour.
For instance, LLMs are able to generate and refine executable reward functions with human feedback~\citep{Kwon2023-rewardmodels,xie2023text2reward}, as well as identify productive behaviour for imitation~\citep{colas2023AugmentingAutotelicAgents,zhongPolicyImprovementUsing}.

We conduct a comprehensive study on how well LLMs and VLMs can provide accurate feedback in grounded environments.
Given a LLM/VLM-based Feedback Model (FM), a language instruction, observation from the environment, and agent behaviour, we compare FM feedback on the agent behaviour to ground truth feedback derived from an optimal policy.
We evaluate state-of-the-art FMs across model sizes, reasoning capabilities, multiple domains, observation modality, and \inferenceconditions.
First, we consider feedback types commonly used in planning: \textbf{binary feedback} on whether the action is optimal; \textbf{preference} on which action out of a pair results in faster task completion; \textbf{action advising} to choose the correct action; \textbf{goal advising} to identify the next subgoal; and \textbf{delta action} to provide action correction.
Second, we consider a diverse selection of domains ranging from symbolic games, verbalized text environments, and continuous control of a robotic arm.
Third, we evaluate how the quality of feedback changes under \inferenceconditions~that affect model performance, such as observation modality, hints about domain dynamics, in-context learning, and chain-of-thought.

We find that Feedback Models can achieve high feedback accuracy across domains.
First, feedback quality improves with larger and reasoning models.
The latter exhibit less class and positional bias, and benefit the most from test-time techniques such as hints about domain dynamics, in-context learning, and chain-of-thought.
Second, out of the feedback types we consider, FMs tend to provide high-quality binary and preference feedback more easily --- this is especially true for non-reasoning models.
Third, feedback quality degrades in domains with complex, hierarchical dynamics, or domains with continuous state and action spaces.
On predominantly visual environments, we find that VLMs consistently outperform LLMs on preference and binary feedback in cases such as continuous control, where image observations are much easier to reason about than textual descriptions of state and action.

To the best of our knowledge, this is the first comprehensive analysis on how well LLMs and VLMs provide across diverse feedback types, domains, and \inferenceconditions.
We show that language model feedback is a promising means to extract learning signals for planning in a diverse selection of environments, without hand-engineered reward functions or human demonstrations.
Future research should investigate how to effectively elicit robust feedback, as well as how to learn from such feedback efficiently.
Given the success of high-quality language model feedback across diverse domains, we advocate for further research into how this feedback can be used to improve policy learning efficiency and effectiveness.
Additionally, we advocate for further research in how to extract high-quality feedback in domains with complex dynamics and continuous state and action spaces --- features that make it particularly challenging to design rewards and label demonstrations.

\section{Methodology}
\label{sec:methodology}

\begin{figure}[!t]
  \vspace{-0.1in}
  \includegraphics[width=\linewidth]{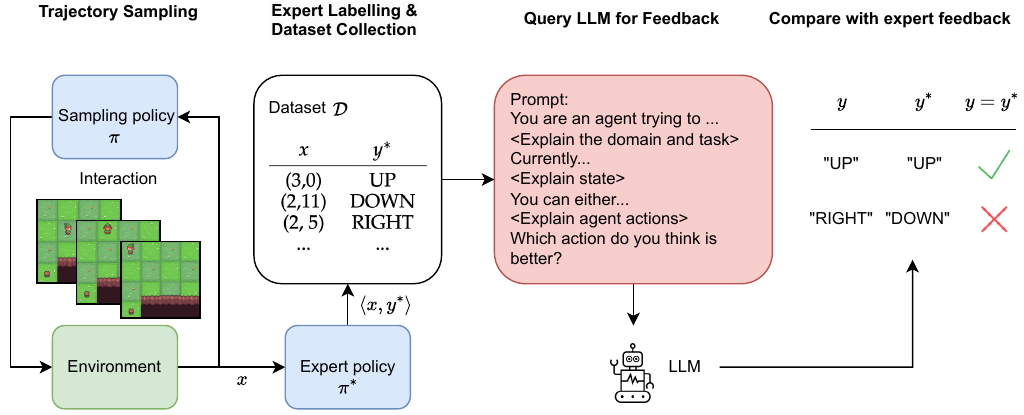}
  \caption{Evaluation of action advising feedback as an example. We sample trajectories according to policy $\pi$, then combine snapshots $x$ from trajectory with expert policy $\pi^*$ to compute ground truth feedback $y^*$.
  For each $(x, y^*)$, we prompt the Feedback Model and compare its response $y$ to the ground truth feedback $y^*$.
  }
\label{fig:pipeline}
\vspace{-0.1in}
\end{figure}

We evaluate how well LLMs and VLMs (which we will refer to as LMs) provide feedback in grounded environments.
Specifically, an LM serves as a~\textbf{Feedback Model} (FM) that provides judgment on how well an agent is performing a task given the task instruction and snapshots of the agent's behaviour.
Figure~\ref{fig:pipeline} presents an overview of how we evaluate the Feedback Model.

We formalize feedback collection using terminology from sequential decision-making. 
Let $S$ denote the state space, $A$ the action space, $T: S\times A \times S \rightarrow [0,1] $ the transition function, $R: S\times A \times S \rightarrow \mathbb{R}$ the reward function, and $\gamma$ the discount factor.
We assume a Markov Decision Process $\mathcal{M}=\langle S, A, T, R, \gamma\ \rangle$, where an agent acts according to policy $\pi_\theta:S \rightarrow A$.
In planning, we find an optimal policy $\pi^*$ that maximizes expected cumulative discounted future reward $G=\mathbb{E}_{(s, a) \sim \pi}[\sum_{t}\gamma^t R(s_t,a_t, s_{t+1})]$.
To characterize how well a state maximizes $G$, we construct the value function $V_\pi(s) = \mathbb{E}_{(s, a) \sim \pi}[G| s_0=s]$ and the state-action value function $Q_\pi(s,a)=\mathbb{E}_{(s, a) \sim \pi}[G| s_0=s, a_0=a]$.
In order to evaluate the Feedback Model, we assume access to an optimal policy $\pi_\theta^*$.
The Feedback Model is given $x$, a \textbf{snapshot} consisting of policy behaviour and environment observations.
The FM then provides $\feedbackmodel(x) = y$, its \textbf{feedback} on the behaviour $x$.
We evaluate the correctness of the judgment by comparing it with \textbf{ground truth feedback} $y^*$ computed using the optimal policy.

\subsection{Types of feedback}
We consider five types of feedback commonly used in planning.
For each, we describe the composition of snapshot $x$ and how the ground truth feedback $y^*$ is computed.

\paragraph{Binary feedback.}
The FM observes a snapshot containing the state and an action $x = (s, a)$ and provides a binary judgment $\feedbackmodel(x) = \ybinary \in \{-1, 1\}$ representing whether the action $a$ is optimal in state $s$.
The ground truth binary feedback is 1 if $\pi^*(s) = a$ and -1 otherwise.
For cases where there exist several optimal actions, we consider $\ybinary$ to be correct if $a$ is any of the optimal actions.
Appendix~\ref{app:domain_dataset} details how we compute $y^*$ for different domains.
Binary feedback estimates binarized advantage value $A_\pi(s,a)=Q_\pi(s,a)-V_\pi(s)$.
\citet{perez2020interactive} and \citet{warnell2018deep} use binary feedback to directly learn policies.
\citet{zhongPolicyImprovementUsing} use binary feedback to identify productive behaviour for imitation learning.

\paragraph{Action advising.}
The FM directly provides the optimal action given the state presented in the snapshot $x = s$.
The ground truth feedback is the output of the optimal policy $\yaction^* = \pi^*(s)$.
For cases where there exist several optimal actions, we consider $\yaction$ to be correct if it is any of the optimal actions. 
LLM has been used to predict optimal actions for long-horizon games~\citep{wangVoyagerOpenEndedEmbodied2023}, robotic control~\citep{saycan2022arxiv}, and computer use~\citep{xie2023text2reward}.

\paragraph{Preference feedback.}
The FM observes a snapshot consisting of the state and a pair of potential actions $x = (s, a_1, a_2)$, and chooses the action more conducive for solving the task (e.g.~higher value $Q(s, a)$).
The ground truth feedback uses $Q^*$ under the optimal policy in order to compute the ground truth preference:
\[
\ypref^* =
\begin{cases} 
+1, & \text{if } \  Q_{\pi^*}(s, a_1) > Q_{\pi^*}(s, a_2) \\
-1, & \text{if }  \ Q_{\pi^*}(s, a_1) < Q_{\pi^*}(s, a_2)  \\
0 & \text{otherwise}
\end{cases}
\]
\citet{lee2021pebble}~use human preference to learn a reward model for reinforcement learning.
\citet{yang2023boosting}~also uses human preferences to learn a preference function, with which they pseudo label data for policy improvement.

\paragraph{Goal advising.}
Goal advising is similar to action advising in that the FM is given a snapshot consisting of the state $x=s$.
However, instead of predicting the next action, the FM predicts the ideal next state $s^\prime$ to best solve the task (e.g.~higher value $V(s^\prime)$.
The ground truth feedback is computed using the value function under the optimal policy $V_{\pi^*}$.
\[
\ygoal = {\mathrm{argmax}}_{s' \in {\mathrm{Reachable}}(s)} V_{\pi^*}(s')  
\]
Here, we use the transition function $T$ to define $\mathrm{Reachable}: s \rightarrow \{s' \in S \mid \exists a \in A, T(s,s',a) > 0 \}$ as a mapping from a state $s$ to a set of states with non-zero transition probability from $s$.
If there are multiple optimal next states, we consider $\ygoal$ to be correct if it is any of the optimal next states.
\citet{Du2023-guidingmodels} use LLMs to generate subgoals and give intermediate rewards to the agent when it achieves the generated subgoals.
\citet{zhao-etal-2024-epo} similarly decompose goals using LLMs, however the low-level goals are then further given to policy LLMs to predict the appropriate action.

\paragraph{Delta action feedback.}
The FM is given a snapshot consisting of the state and an action $x = (s, a)$, and provides a corrective delta action $\Delta$ such that $a + \Delta$ is more conducive for solving the task than $a$ (e.g.higher value~$Q(s, a+\Delta) > Q(s, a)$).
The ground truth feedback uses the optimal policy in order to compute the ground truth delta action $\ydelta^* = \pi^*(s) - a$.
We consider $\ydelta$ to be correct if it is within some threshold $\epsilon$ of the ground truth $\ydelta^*$.
\citet{liuInteractiveRobotLearning2023} map verbal feedback provided by human observers to a set of delta actions, which are used to correct actions produced by a policy.
\citet{colas2023AugmentingAutotelicAgents} convert agent trajectories to language descriptions, and use LLMs to retroactively provide corrections to actions in the trajectory.

\begin{figure}[!t]
  \vspace{-0.15in}
\begin{subfigure}[]{0.32\textwidth}
  \includegraphics[width=\linewidth]{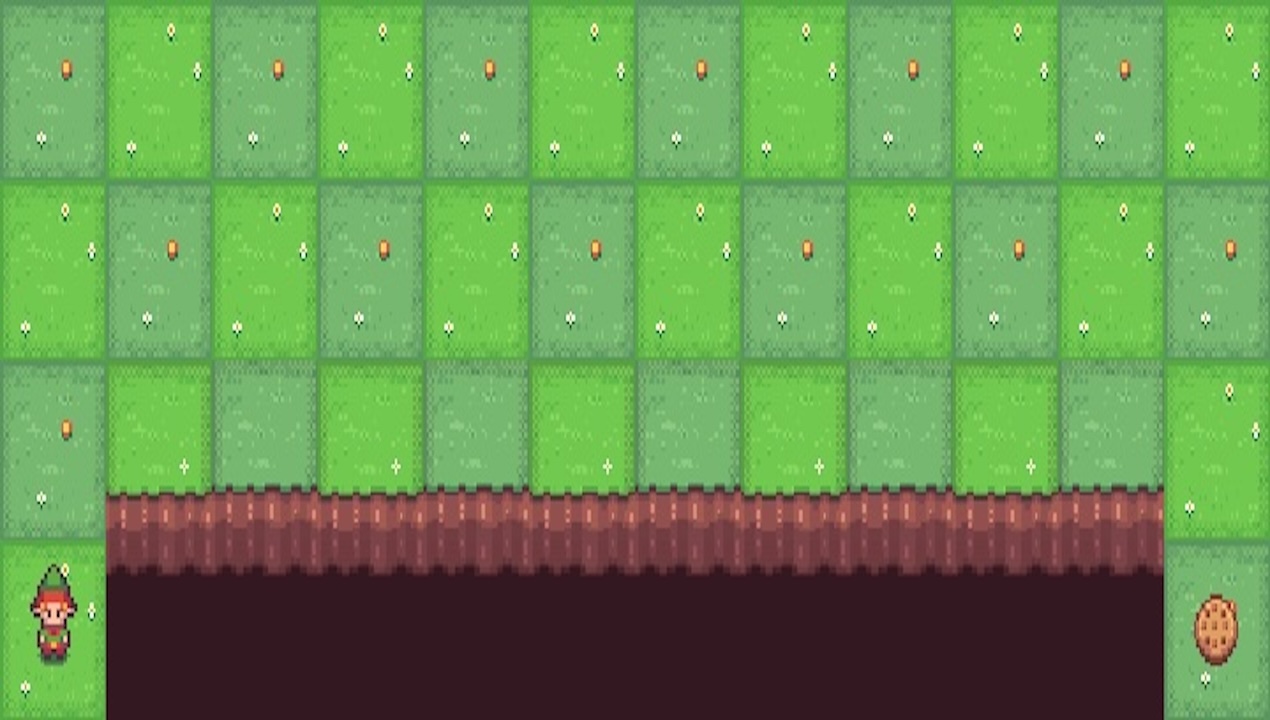}
  \caption{Cliff Walking}\label{fig:ciffwalking}
\end{subfigure}
\begin{subfigure}[]{0.32\textwidth}%
  \includegraphics[width=\linewidth]{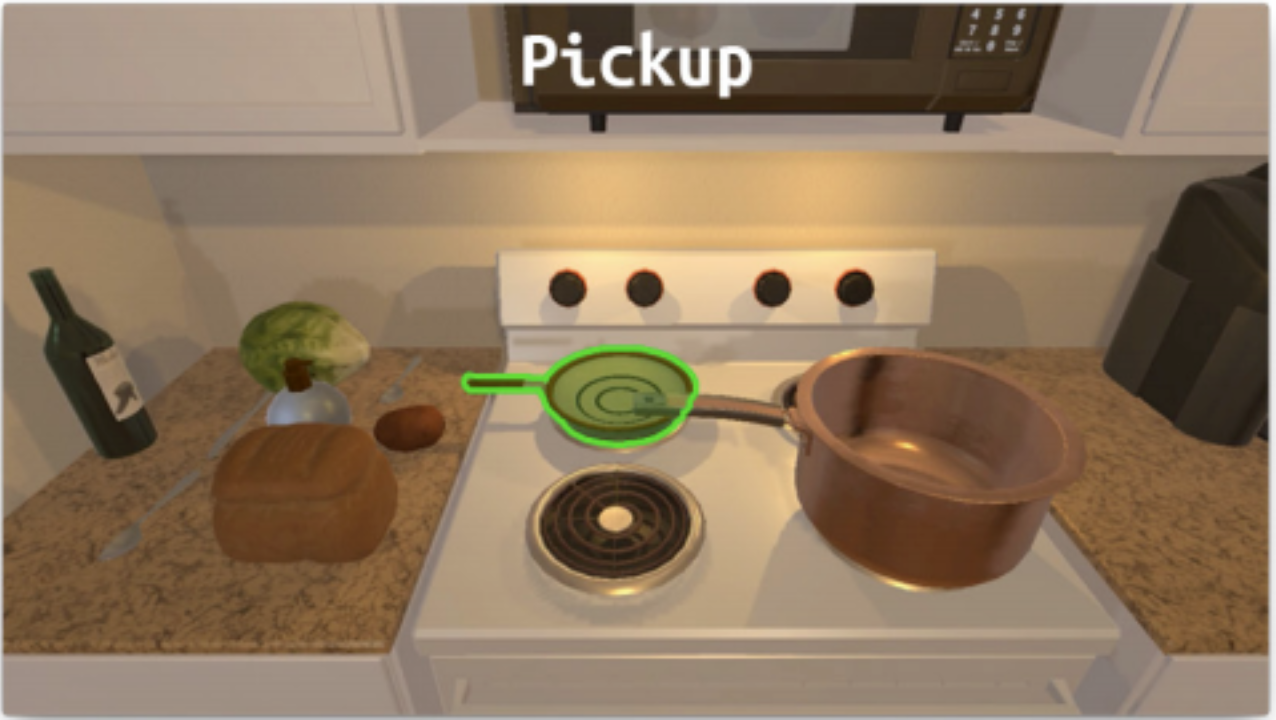}
  \caption{ALFWorld}\label{fig:alf}
\end{subfigure}
\begin{subfigure}[]{0.32\textwidth}%
  \includegraphics[width=\linewidth]{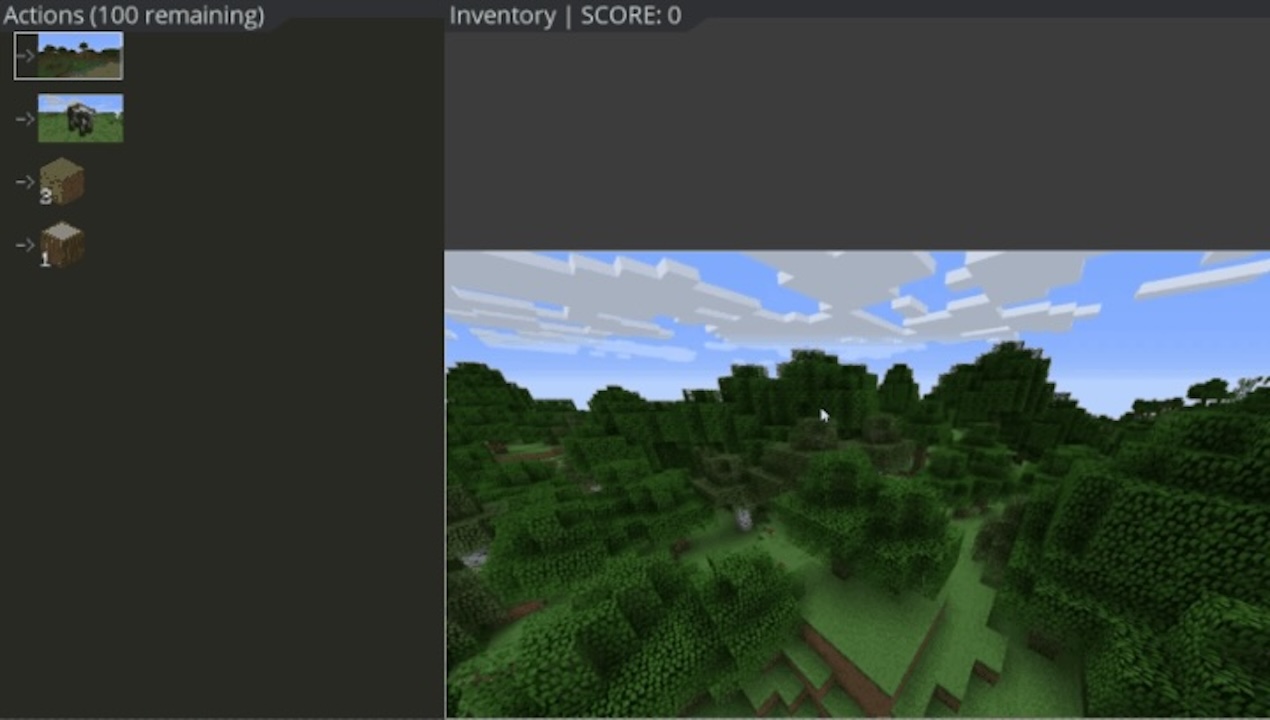}
  \caption{HierarchyCraft}\label{fig:minecraft}
\end{subfigure}
\begin{subfigure}[]{0.32\textwidth}
  \includegraphics[width=\linewidth]{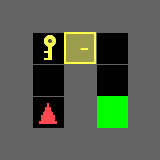}
  \caption{ MiniGrid Door Key}\label{fig:doorkey}
\end{subfigure} \ \ 
\begin{subfigure}[]{0.32\textwidth}%
  \includegraphics[width=\linewidth]{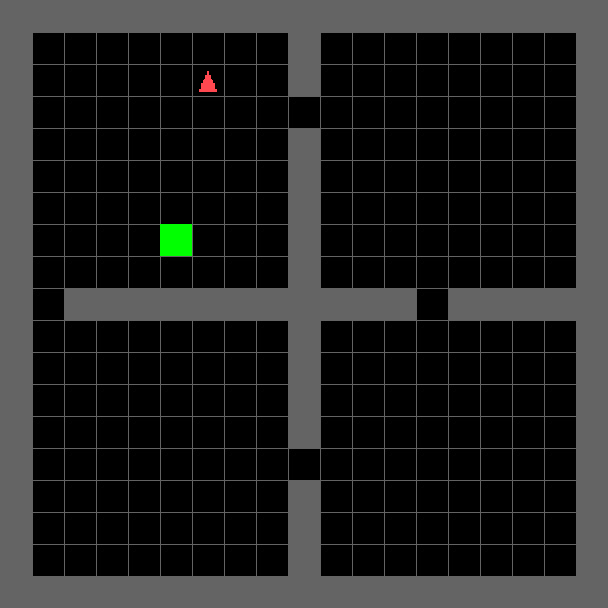}
  \caption{MiniGrid Four Rooms}\label{fig:fourrooms}
\end{subfigure} \ \ 
\begin{subfigure}[]{0.32\textwidth}%
  \includegraphics[width=\linewidth]{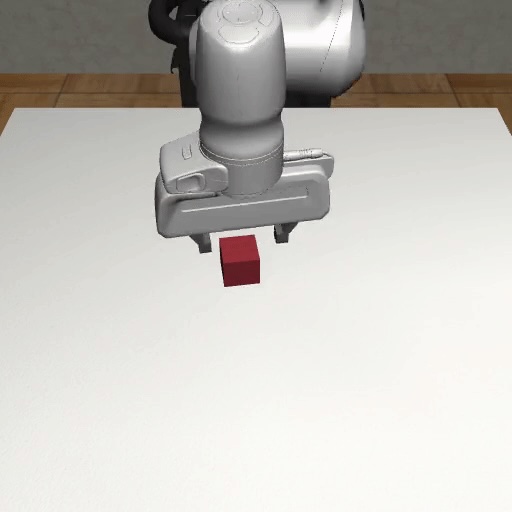}
  \caption{Robomimic Lift Object}\label{fig:robomimic}
\end{subfigure}
\caption{Six evaluation domains for language model feedback.}
\label{fig:domains}
    \vspace{-0.1in}
\end{figure}

\subsection{Task Domains}

In order to measure how well Feedback Models provide feedback across diverse settings, we collect policy behaviours from several distinct domains that vary in observation space, action space, degree of observability, and task settings.
Figure~\ref{fig:domains} illustrates these domains, which include symbolic games CliffWalking~\citep{sutton2018reinforcement} and MiniGrid~\citep{MinigridMiniworld23}, navigation and high-level manipulation text simulators ALFWorld~\citep{ALFWorld20} and HierchyCraft~\citep{HierarchyCraft}, and continuous control Robomimic~\citep{robomimic2021}.
We evaluate LLM feedback for all domains, and VLM feedback for predominantly visual domains (MiniGrid Door Key and Robomimic).
Appendix~~\ref{app:prompt_examples} shows examples of FM prompts for each environment.

\paragraph{Cliff Walking}
The agent navigates to a fixed goal location in a grid of size 4x12 without falling off the cliff located in the last row.
The agent has full observability of the grid, along with its current position (e.g.~\texttt{(3, 0)}), and chooses among the actions~\texttt{right},~\texttt{left},~\texttt{up}, and~\texttt{down} to navigate to the goal.

\paragraph{MiniGrid~\citep{MinigridMiniworld23}}
We consider two tasks where the agent navigates between rooms in order to reach a goal location.
In Four Rooms (Fig~\ref{fig:fourrooms}), the agent needs to navigate four rooms to reach the goal.
It chooses among the actions~\texttt{turn left},~\texttt{turn right},~\texttt{move forward}.
In Door Key (Fig~\ref{fig:doorkey}), the agent must first pick up the key, then unlock the door, then reach the goal.
The agent chooses among actions~\texttt{turn left},~\texttt{turn right},~\texttt{move forward},~\texttt{pick up}, and~\texttt{unlock door}.
For both tasks, the agent has full observability of the room layout and content.
We evaluate VLM feedback on Minigrid Door Key, as it involves more visual understanding components like locked door, unlocked door, wall and key.
Here, we use a screenshot of the current grid as the visual observation.

\paragraph{HierarchyCraft~\citep{HierarchyCraft}}
We consider the task of finding a diamond.
The agent must navigate to appropriate locations to collect materials (e.g.~iron, wood) and craft items (e.g.~iron pick axe) helpful for procuring the diamond.
The action space consists of 201 discrete actions (e.g.~searching for wood in the forest).
The agent observes its inventory and the items in the environment via text description.
This environment is partially observable, as objects in a location are hidden by default.

\paragraph{ALFWorld~\citep{ALFWorld20}}
The agent navigates and performs high-level interactions with objects in a house in order to accomplish high-level tasks (e.g.~wash mugs and put them in the cabinet).
The agent observes the goal and text descriptions of the scene, and chooses among a list of valid actions described in language.
This environment is partially observable, as objects in a location are hidden by default (e.g.~cups in a cabinet).

\paragraph{Robomimic~\citep{robomimic2021}}
We consider the lift object task, where the agent controls a 7-DoF robot arm to lift an object.
The agent has full observability of the end effector position, the object position.
It chooses from a list of control vectors that move the end effector and adjust the gripper state.
Similar to~\citet{robomimic2021}, we assume an external policy that can accurately control the gripper rotation.
In our experiments, we represent the observations via text description and provide a list of delta actions for the agent to choose from.
For VLM feedback, we use a single image instead of a text description of the current observation.
The image is captured from a camera in front of the desk and robot arm, as shown in Figure~\ref{fig:robomimic}.

\begin{wraptable}[]{r}{0.5\linewidth}

 \vspace{-0.45\baselineskip}
  \centering
  \begin{tabular}{@{}lccc@{}}
    \toprule
    Domain & Pref & Act & Binary \\
    \midrule
    Cliff Walking & 444 & 37 & 148 \\
    Door Key & 2000 & 100 & 500 \\
    Four Rooms & 3000 & 500 & 1500 \\
    Robomimic (Lift) & 2015 & 2015 & 2015 \\
    ALFWorld & 2778k & 3942 & 105k \\
    MineCraft & 139k & 1000 & 11k \\
    \bottomrule
  \end{tabular}
  \caption{Dataset size for each domain and feedback type.
  The listed types are preference feedback, action advising, and binary feedback.
  Delta action is supported in Robomimic with 2015 examples.
  Goal advising is supported in Cliff Walking with 37 examples.}
  \label{tab:dataset_size}
\end{wraptable}

\subsection{Dataset Collection}
For each domain, we collect a dataset of policy behaviour $\mathcal{D} = \{x_1, x_2 \ldots\}$ by sampling from a policy $\pi$.
For each snapshot $x_i$, we first compute the set of optimal actions using the optimal policy $\pi^*$, then compute the ground truth feedback $y^*$.

For Cliff Walking, we enumerate all states/ actions for a dataset of 37 snapshots for action/goal advising, 148 for binary feedback, and 444 for preference feedback.
For other domains, we use 3 choices of $\pi$: expert $\pi = \pi^*$, random policy that uniformly samples actions, and half-expert which alternates between the two with 50\% chance.
For each state, we enumerate all actions to construct $\mathcal{D}$.
For example, for Four rooms, we sample $N$ states $s_1 \ldots s_N$ from $\pi$ and create a state-action pair for each of the 3 actions to build a binary feedback dataset of $3N$ examples.

We cannot compute the optimal actions for Robomimic Lift Object, as it is impossible to enumerate all possible continuous actions.
Following~\citet{robomimic2021}, we instead discretize the action space using a threshold based on empirical evaluation of a behavioural cloning expert, and use this threshold to enumerate the set of delta actions.
The size of the collected dataset is shown in Table~\ref{tab:dataset_size}.

\subsection{Feedback Conditions}
Across each domain, we evaluate FM \inferenceconditions~that are known to affect LLM performance in grounded environments.

\paragraph{In-Context Learning (ICL)}
ICL improves LLM performance by providing input and output examples~\citep{brown2020language}, from which LLMs deduce patterns that help generalize to new tasks~\citep{dong2022survey,chen2024icleval}.
In this setting, we combine instructions, observations, and ground truth feedback as ICL examples for the FM.

\paragraph{Chain-of-Thought and Thinking Guides (CoT)}
CoT techniques encourage LLMs to generate intermediate steps towards the final solution~\citep{wei2022chain}, and consistently improve performance for reasoning problems such as program synthesis~\citep{chen2022program} and planning~\citep{yao2023tree}.
We incorporate CoT in all experiments.
In this additional setting, we design planning-specific questions, called \textbf{Thinking Guides}, for the FM to answer to encourage the completion of helpful high-level intermediate steps (e.g.~reasoning about its inventory).
Instead of trying to deduce the reasoning procedure from ICL examples, Thinking Guides directly gives FMs high-level hints for how to reason.

\paragraph{Access to High-level Domain Dynamics}
\citet{kaiser2019model} and~\citet{hafner2023mastering} show that knowledge of environment dynamics improve model-based reinforcement learning.
In this setting, we enumerate the transition dynamics of the environment and either provide a full description of the dynamics or a limited partial version explaining only legal actions.
We additionally explore a second setting, where instead of directly providing environment dynamics, we provide~\textbf{history context} to the FM model.
This evaluates the FM's ability to infer relevant dynamics from the history.
In order to provide history context for Cliff Walking, we use an expert policy to sample trajectories.

\begin{table}[!t]
    \centering
    \begin{tabularx}{\linewidth}{lXXXX}
        \toprule
        Domain & Preference & Action Adv & Bin Feedback & Goal Adv\\
        \midrule
        Cliff Walking &
            \coloredmatrix{0.33}{0.77}{0.91}{1.00}
                    &
            \coloredmatrix{0.68}{0.73}{0.99}{1.00}
                    &
            \coloredmatrix{0.61}{0.73}{1.00}{0.98}
                    &
            \coloredmatrix{0.37}{0.77}{1.00}{1.00}
                    \\
                    \midrule
    Door Key &
            \coloredmatrix{0.66}{0.75}{0.89}{0.98}
                    &
            \coloredmatrix{0.44}{0.44}{0.81}{0.99}
                    &
            \coloredmatrix{0.78}{0.75}{0.92}{1.00}
                    & \notapplicable
                    \\
                    \midrule
    Four Rooms &
            \coloredmatrix{0.59}{0.66}{0.65}{0.75}
                    &
            \coloredmatrix{0.32}{0.42}{0.61}{0.78}
                    &
            \coloredmatrix{0.59}{0.69}{0.66}{0.81}
                    & \notapplicable
                    \\
                    \midrule
    Robomimic &
            \coloredmatrix{0.55}{0.67}{0.55}{0.56}
                    &
            \coloredmatrix{0.01}{0.00}{0.00}{0.00}
                    &
            \coloredmatrix{0.47}{0.50}{0.51}{0.46}
                    & \notapplicable
                    \\
                    \midrule
    ALFWorld &
            \coloredmatrix{0.49}{0.67}{0.78}{0.83}
                    &
            \coloredmatrix{0.60}{0.83}{0.85}{0.83}
                    &
            \coloredmatrix{0.76}{0.74}{0.77}{0.76}
                    & \notapplicable
                    \\
                    \midrule
    HierachyCraft &
            \coloredmatrix{0.51}{0.51}{0.81}{0.89}
                    &
            \coloredmatrix{0.58}{0.61}{0.65}{0.78}
                    &
            \coloredmatrix{0.81}{0.85}{0.88}{0.91}
                    & \notapplicable
                    \\
        \bottomrule
    \end{tabularx}
    \caption{
    Feedback accuracy for \colorbox{red!30}{Llama 3.1 8B}, \colorbox{blue!30}{Llama 3.1 70B}, \colorbox{green!30}{QwQ 32B}, and \colorbox{yellow!30}{R1}.
    Delta action is only applicable for continuous state spaces (Robomimic), its feedback column is not shown due to space constraint --- Robomimic Delta Action numbers are 0.07 for Llama 3.1 8B, 0.12 for Llama 3.1 70B, 0.03 for QwQ, and 0.07 for R1.
    For goal advising, we only consider cases ground truth feedback is computable --- there exists an intermediate state that the FM can propose which the optimal policy must encounter (Cliff Walking).
    }
    \label{tab:all_normal_expert}
    \vspace{-0.1in}
\end{table}

\section{Experiments and Analyses}
We design experiments to quantify how well LMs provide feedback across different feedback types, environment domains, observation modality, and \inferenceconditions.
We conduct our study with LLama 3.1 7B, Llama 3.1 70B~\citep{grattafiori2024llama3herdmodels}, QwQ 32B~\citep{qwq-32b-preview}, and DeepSeek R1~\citep{deepseekai2025deepseekr1incentivizingreasoningcapability}.
The Llama models are non-reasoning models, while QwQ and R1 are reasoning models with performance similar to the best closed-source models.
We evaluate VLM feedback with Qwen 2.5 VL 7B, Qwen 2.5 VL 32B, Qwen 2.5 VL 72B~\citep{bai2025qwen2} and GPT o4 mini.
We show Feedback Model prompts in Appendix~\ref{app:prompt_examples} for each feedback type and for each feedback condition.
We consider malformatted feedback and feedback with illegal values to be incorrect. 
During the evaluation, we down-sample the large datasets for ALFWorld and Robomimic to 1000.
For preference feedback, we only evaluate feedback on pairs of optimal and non-optimal actions, as opposed to pairs of equally optimal actions.

The accuracy of Feedback Models across domains and feedback types under the expert policy is shown in Table~\ref{tab:all_normal_expert}.
We show results on snapshots from expert policy, except for the smaller Cliff Walking domain where we enumerate all possible states.
Additional results using random policies and mixed policies are shown in Appendix~\ref{app:otherpolicies}.

\paragraph{LLMs can provide high-quality feedback} 
Table~\ref{tab:all_normal_expert} shows that LLMs can provide accurate feedback across domains.
Moreover, larger models and reasoning models provide more accurate feedback regardless of feedback type, across different domains.
For instance, the smaller QwQ consistently outperforms the larger non-reasoning Llama 3.1 70B.
One source of small FM errors is that small FMs are prone to bias.
First, Appendix~\ref{app:f1scores} show that they are biased towards predicting actions as non-optimal, resulting in much worse recall than precision for binary feedback.
Second, Appendix~\ref{app:positinalbias} show that they are biased towards preferring the first action, resulting in much worse accuracy when the ground truth action is located second for preference feedback.
In contrast, larger FMs exhibit far less positive/negative bias in binary feedback and positional bias in preference feedback. 
\begin{table}[t]
    \centering
    \begin{tabularx}{\linewidth}{lXXXX}
        \toprule
        Domain & Preference & Action Adv & Bin Feedback  & Goal Adv\\
        \midrule
        
        Cliff Walking
                    & \coloredmatrixdelta{0.59}{\up{.26}}{0.78}{\up{.01}}{0.95}{\up{.04}}{0.97}{\down{.03}}
                    & 
                    \coloredmatrixdelta{0.57}{\down{.15}}{0.62}{\down{.11}}{1.00}{\up{.01}}{0.97}{\down{.03}}
                    &
                    \coloredmatrixdelta{0.69}{\up{.08}}{0.80}{\up{.07}}{0.95}{\down{.05}}{0.98}{\same}
                    &
                    \coloredmatrixdelta{0.58}{\up{.21}}{0.97}{\up{.20}}{0.97}{\down{.03}}{0.98}{\down{.02}}
                    \\
                    \midrule
                Door Key    &
                    \coloredmatrixdelta{0.50}{\down{.11}}{0.71}{\down{.04}}{0.94}{\up{.05}}{0.98}{\same}
                    &
                    \coloredmatrixdelta{0.35}{\down{.09}}{0.55}{\up{.11}}{0.87}{\up{.06}}{0.98}{\down{.01}}
                    &
                    \coloredmatrixdelta{0.68}{\down{.10}}{0.72}{\down{.03}}{0.92}{\same}{0.98}{\down{.02}}
                    & \notapplicable
                    \\
                    \midrule

        Four Rooms  &
                    \coloredmatrixdelta{0.58}{\down{.01}}{0.61}{\down{.05}}{0.65}{\same}{0.79}{\up{.04}}
                    &
                    \coloredmatrixdelta{0.53}{\up{.20}}{0.54}{\up{.12}}{0.60}{\down{.01}}{0.83}{\up{.05}}
                    &
                    \coloredmatrixdelta{0.42}{\down{.17}}{0.50}{\down{.19}}{0.54}{\down{.11}}{0.86}{\up{.05}}
                        & \notapplicable
                        \\
                        \midrule
        ALFWorld    &
                    \coloredmatrixdelta{0.50}{\up{.01}}{0.71}{\up{.09}}{0.86}{\up{.08}}{0.85}{\up{.02}}
                    &
                    \coloredmatrixdelta{0.69}{\up{.09}}{0.85}{\up{.01}}{0.88}{\up{.03}}{0.93}{\up{.10}}
                    &
                    \coloredmatrixdelta{0.74}{\down{.02}}{0.77}{\up{.03}}{0.76}{\down{.01}}{0.77}{\up{.01}}
                    & \notapplicable
                    \\
                    \midrule
            HierachyCraft
                    & 
                    \coloredmatrixdelta{0.56}{\up{.05}}{0.93}{\up{.42}}{0.89}{
                    \up{.08}}{0.93}{\up{.04}}
                    &
                    \coloredmatrixdelta{0.63}{\up{.05}}{0.68}{\up{.07}}{0.52}{\down{.13}}{0.84}{\up{.06}}
                    &
                    \coloredmatrixdelta{0.83}{\up{.02}}{0.87}{\up{.02}}{0.87}{\down{.01}}{0.89}{\down{.02}}
                    &
                    \notapplicable \\
            \bottomrule
        \end{tabularx}
    \caption{Feedback accuracy with in-context learning, the arrows show accuracy change with this condition relative to baseline}
    \label{tab:all_icl_expert}
\end{table}

\begin{table}[t]
    \centering
    \begin{tabularx}{\linewidth}{lXXXX}
        \toprule
        Domain & Preference & Action Adv & Bin Feedback  & Goal Adv\\
        \midrule        
        Cliff Walking
                    & \coloredmatrixdelta{0.49}{\up{.16}}{0.47}{\down{.30}}{0.99}{\up{.08}}{1.00}{\same}
                    & 
                        \coloredmatrixdelta{0.68}{\same}{0.92}{\up{.19}}{0.99}{\same}{1.00}{\same} &
                        \coloredmatrixdelta{0.52}{\down{0.09}}{0.50}{\down{.23}}{1.00}{\same}{0.93}{\down{.05}} &
                        \coloredmatrixdelta{0.57}{\up{.20}}{0.92}{\up{.15}}{1.00}{\same}{0.97}{\down{.03}}
                    \\
                    \midrule
        Door Key    &
                        \coloredmatrixdelta{0.53}{\down{.13}}{0.72}{\down{.03}}{0.94}{\up{.05}}{0.98}{\same} &
                        \coloredmatrixdelta{0.47}{\up{.03}}{0.41}{\down{.03}}{0.89}{\up{.07}}{0.99}{\same} &
                        \coloredmatrixdelta{0.78}{\same}{0.76}{\up{.01}}{0.94}{\up{.02}}{1.00}{\same} &
                        \notapplicable 
                    \\
                    \midrule

        Four Rooms  &
                        \coloredmatrixdelta{0.53}{\down{.03}}{0.58}{\down{.08}}{0.71}{\up{.06}}{0.81}{\up{.06}} &
                        \coloredmatrixdelta{0.47}{\up{.15}}{0.43}{\up{.01}}{0.59}{\down{.02}}{0.75}{\down{.03}} &
                        \coloredmatrixdelta{0.63}{\up{.04}}{0.62}{\down{.07}}{0.71}{\up{.05}}{0.81}{\same} &
                        \notapplicable \\
                        \midrule
        ALFWorld    &
                        \coloredmatrixdelta{0.51}{\up{.02}}{0.68}{\up{.01}}{0.82}{\up{.03}}{0.88}{\up{.05}} &
                        \coloredmatrixdelta{0.59}{\down{.01}}{0.83}{\same}{0.83}{\down{.02}}{0.83}{\same} &
                        \coloredmatrixdelta{0.72}{\down{.04}}{0.76}{\down{.02}}{0.76}{\down{.01}}{0.76}{\same} &
                        \notapplicable \\
                        \midrule
            HierachyCraft
                    & 
                        \coloredmatrixdelta{0.51}{\same}{0.78}{\up{.27}}{0.84}{\up{.03}}{0.89}{\same} &
                        \coloredmatrixdelta{0.37}{\down{.21}}{0.63}{\up{.02}}{0.73}{\up{.08}}{0.78}{\same} &
                        \coloredmatrixdelta{0.79}{\down{.02}}{0.85}{\same}{0.88}{\same}{0.90}{\down{.01}} &
                        \notapplicable \\
            \bottomrule
        \end{tabularx}
    \caption{Feedback accuracy with Thinking Guides, the arrows show accuracy change with this condition relative to baseline.}
    \label{tab:thinking_guides}
    \vspace{-0.1in}
\end{table}

\begin{table}[!t]
    \centering
    \begin{tabularx}{\linewidth}{lXXXX}
        \toprule
        Domain & Preference & Action Adv & Bin Feedback & Delta Action\\
        \midrule
    Door Key &
            \coloredmatrix{0.39}{0.52}{0.60}{0.61}
                    &
            \coloredmatrix{0.42}{0.27}{0.32}{0.43}
                    &
            \coloredmatrix{0.48}{0.65}{0.64}{0.77}
                    & \notapplicable
                    \\
    In Context Learning    &
                    \coloredmatrixdelta{0.40}{\up{.01}}{0.53}{\up{.01}}{0.58}{\down{.02}}{0.58}{\down{.03}}
                    &
                    \coloredmatrixdelta{0.47}{\up{.05}}{0.18}{\down{.09}}{0.24}{\down{.08}}{0.37}{\down{.06}}
                    &
                    \coloredmatrixdelta{0.74}{\up{.22}}{0.75}{\up{.10}}{0.67}{\up{.03}}{0.70}{\down{.07}}
                    & \notapplicable
                    \\
    Thinking Guides    &
            \coloredmatrixdelta{0.39}{\same}{0.49}{\down{.03}}{0.45}{\down{.15}}{0.60}{\down{.01}}
            &
            \coloredmatrixdelta{0.51}{\up{.09}}{0.18}{\down{.09}}{0.23}{\down{.09}}{0.28}{\down{.15}}
            &
            \coloredmatrixdelta{0.40}{\down{.08}}{0.31}{\down{.34}}{0.42}{\down{.22}}{0.48}{\down{.29}}
            & \notapplicable
            \\
    Unk & 
                        \coloredmatrixdelta{0.50}{\up{.11}}{0.44}{\down{.08}}{0.54}{\down{.06}}{0.45}{\down{.16}} &
                        \coloredmatrixdelta{0.27}{\down{.15}}{0.15}{\down{.12}}{0.26}{\down{.06}}{0.15}{\down{.28}} &
                        \coloredmatrixdelta{0.54}{\up{.06}}{0.51}{\down{.14}}{0.58}{\down{.06}}{0.57}{\down{.20}} &

                        \notapplicable \\
            Unk, History & 
                        \coloredmatrixdelta{0.50}{\up{.11}}{0.71}{\up{.19}}{0.66}{\up{.06}}{0.55}{\down{.06}} &
                        \coloredmatrixdelta{0.40}{\down{.02}}{0.50}{\up{.23}}{0.40}{\up{.08}}{0.42}{\down{.01}} &
                        \coloredmatrixdelta{0.74}{\up{.26}}{0.67}{\up{.02}}{0.67}{\up{.03}}{0.67}{\down{.10}} &
                        \notapplicable \\

            History & 
                        \coloredmatrixdelta{0.22}{\down{.17}}{0.72}{\up{.20}}{0.72}{\up{.12}}{0.61}{\same} &
                        \coloredmatrixdelta{0.49}{\up{.07}}{0.59}{\up{.22}}{0.47}{\up{.15}}{0.49}{\up{.06}} &
                        \coloredmatrixdelta{0.72}{\up{.24}}{0.75}{\up{.10}}{0.74}{\up{.10}}{0.79}{\up{.02}} &

                        \notapplicable
                    \\
                    \midrule
    Robomimic &
            \coloredmatrix{0.82}{0.77}{0.82}{0.70}
                    &
            \coloredmatrix{0.00}{0.00}{0.00}{0.00}
                    &
            \coloredmatrix{0.43}{0.18}{0.04}{0.09}
                    & 
            \coloredmatrix{0.01}{0.05}{0.04}{0.01}
                    \\
    History & 
                        \coloredmatrixdelta{0.54}{\down{.28}}{0.75}{\down{.02}}{0.80}{\down{.02}}{0.76}{\up{.06}} &
                        \coloredmatrixdelta{0.00}{\same}{0.00}{\same}{0.00}{\same}{0.00}{\same} &
                        \coloredmatrixdelta{0.56}{\up{.13}}{0.62}{\up{.44}}{0.62}{\up{.58}}{0.45}{\up{}} &
                        \coloredmatrixdelta{0.02}{\up{.01}}{0.02}{\down{.03}}{0.03}{\up{.01}}{0.01}{\up{.03}}  \\

        \bottomrule
    \end{tabularx}
    \caption{
    VLM Feedback accuracy for \colorbox{red!30}{Qwen 2.5 VL 7B}, \colorbox{blue!30}{Qwen 2.5 VL 32B}, \colorbox{green!30}{Qwen 2.5 VL 72B}, and \colorbox{yellow!30}{o4-mini}.
    }
    \label{tab:all_normal_expert_image}
    \vspace{-0.1in}
\end{table}
\paragraph{Preference and Binary Feedback are easier to provide}
Preference and Binary Feedback are generally the easiest forms of feedback to accurately provide across domains, especially for non-reasoning models.
For instance, Llama 3.1 8B and 70B provide significantly more accurate binary feedback and preference feedback than action advising.
In contrast, larger models such as QwQ and R1 are able to provide action advising with high quality similar to binary and preference feedback.

\paragraph{Image observation allows better feedback in continuous domains}
VLM feedback results are shown in Table~\ref{tab:all_normal_expert_image}.
For predominantly visual domains, VLMs do not necessarily outperform LLMs if one can verbalize the environment succinctly.
For example, in the Minigrid Door Key, we observed that even one of our best models, Qwen 2.5VL 72B shows worse performance than the model with similar sizes (Llama 3 70B) in text feedback.
However, for Robomimic, in which it is difficult to provide succinct descriptions of states and actions, VLMs give preference up to 82\% accuracy and binary feedback up to 62\% accuracy, outperforming even bigger LLMs (see Table~\ref{tab:all_normal_expert}).
Here, text observation requires a deep understanding of mechanical dynamics, while image observation is more intuitive to understand and hence gives higher accuracy.

\paragraph{Feedback quality degrades significantly in continuous state/action spaces}
Table~\ref{tab:all_normal_expert} shows that in the Robomimic continuous control domain, no model is able to provide accurate feedback.
We observe similar patterns for VLMs, with the exception of preference feedback.
\footnote{OLAF ~\citep{liuInteractiveRobotLearning2023} similarly extract language feedback for Robomimic, however unlike in OLAF, humans do not facilitate querying feedback in our experiments.}
We additionally experiment with egocentric verbalization, which has been shown to improve LLM planning in embodied environments~\citep{ALFWorld20,Schumann2024-velmaview}, however experiments in Appendix~\ref{app:egocentric} show that this does not significantly improve performance of any FMs. 
Our results suggest that current foundation models struggle with interpreting continuous state and action spaces without human prior knowledge.

\begin{table}[t]
    \centering
    \begin{tabularx}{\linewidth}{lXXXX}
        \toprule
        Domain & Preference & Action Adv & Bin Feedback  & Goal Adv\\
        \midrule        
        Cliff Walking Unk &
                        \coloredmatrix{0.49}{0.53}{0.71}{0.85} &
                        \coloredmatrix{0.19}{0.21}{0.38}{0.58} &
                        \coloredmatrix{0.19}{0.39}{0.41}{0.74} &
                        \coloredmatrix{0.11}{0.22}{0.26}{0.52} \\
             Full & 
                        \coloredmatrixdelta{0.51}{\up{.02}}{0.77}{\up{.24}}{0.95}{\up{.24}}{1.00}{\up{.15}} &
                        \coloredmatrixdelta{0.20}{\up{.01}}{0.44}{\up{.23}}{1.00}{\up{.62}}{1.00}{\up{.42}} &
                        \coloredmatrixdelta{0.47}{\up{.28}}{0.39}{\same}{1.00}{\up{.59}}{0.98}{\up{.24}} &
                        \coloredmatrixdelta{0.15}{\up{.04}}{0.69}{\up{.45}}{1.00}{\up{.74}}{1.00}{\up{.48}} \\
            History & 
                        \coloredmatrixdelta{0.48}{\down{.01}}{0.53}{\same}{0.70}{\down{.01}}{0.85}{\same} &
                        \coloredmatrixdelta{0.19}{\same}{0.20}{\down{.01}}{0.47}{\up{.06}}{0.68}{\down{.06}} &
                        \coloredmatrixdelta{0.25}{\up{.06}}{0.41}{\up{.02}}{0.42}{\.01}{0.80}{\up{.06}} &
                        \coloredmatrixdelta{0.25}{\up{.14}}{0.41}{\up{.19}}{0.42}{\up{.16}}{0.84}{\up{.32}} \\
                        \midrule

            Door Key Unk & 
                        \coloredmatrix{0.53}{0.76}{0.91}{0.92} &
                        \coloredmatrix{0.41}{0.51}{0.82}{0.87} &
                        \coloredmatrix{0.79}{0.75}{0.92}{0.94} &
                        \notapplicable \\
            Full & 
                        \coloredmatrixdelta{0.66}{\up{.13}}{0.75}{\down{.01}}{0.89}{\down{.02}}{0.98}{\up{.06}} &
                        \coloredmatrixdelta{0.44}{\up{.03}}{0.44}{\down{.07}}{0.81}{\down{.01}}{0.99}{\up{.12}} &
                        \coloredmatrixdelta{0.78}{\down{.01}}{0.79}{\up{.04}}{0.92}{\same}{1.00}{\up{.06}} &
                        \notapplicable \\
            History & 
                        \coloredmatrixdelta{0.54}{\up{.01}}{0.75}{\down{.01}}{0.89}{\down{.02}}{0.91}{\down{.01}} &
                        \coloredmatrixdelta{0.40}{\down{.04}}{0.50}{\down{.01}}{0.78}{\down{.02}}{0.86}{\down{.01}} &
                        \coloredmatrixdelta{0.79}{\same}{0.75}{\same}{0.90}{\down{.02}}{0.93}{\down{.01}} &
                        \notapplicable \\
            \midrule
            HierachyCraft Unk
                        &
                        \coloredmatrix{0.51}{0.51}{0.81}{0.89}
                        &
                        \coloredmatrix{0.58}{0.61}{0.65}{0.78}
                        &
                        \coloredmatrix{0.81}{0.85}{0.88}{0.91}
                        & \notapplicable \\
            Full  & 
                        \coloredmatrixdelta{0.49}{\down{.02}}{0.83}{\up{.32}}{0.90}{\up{.09}}{0.87}{\down{.02}} &
                        \coloredmatrixdelta{0.30}{\down{.28}}{0.65}{\up{.04}}{0.78}{\up{.13}}{0.80}{\up{.02}} &
                        \coloredmatrixdelta{0.79}{\down{.02}}{0.82}{\down{.03}}{0.92}{\up{.04}}{0.92}{\up{.01}} &
                        \notapplicable \\
            Legal Action & 
                        \coloredmatrixdelta{0.51}{\same}{0.80}{\up{.29}}{0.87}{\up{.06}}{0.92}{\up{.03}} &
                        \coloredmatrixdelta{0.46}{\down{.12}}{0.60}{\down{.01}}{0.65}{\same}{0.80}{\up{.02}} &
                        \coloredmatrixdelta{0.81}{\same}{0.83}{\down{.02}}{0.88}{\same}{0.91}{\same} &
                        \notapplicable \\
            History & 
                        \coloredmatrixdelta{0.52}{\up{.01}}{0.66}{\up{.15}}{0.90}{\up{.09}}{0.91}{\up{.02}} &
                        \coloredmatrixdelta{0.45}{\down{.12}}{0.64}{\up{.03}}{0.84}{\up{.19}}{0.88}{\up{.10}} &
                        \coloredmatrixdelta{0.78}{\down{.03}}{0.84}{\down{.01}}{0.91}{\up{.03}}{0.92}{\up{.01}} &
                        \notapplicable \\
                                    \midrule
            Robomimic &
            \coloredmatrix{0.55}{0.67}{0.55}{0.56}
                    &
            \coloredmatrix{0.01}{0.00}{0.00}{0.00}
                    &
            \coloredmatrix{0.47}{0.50}{0.51}{0.46}
                    &                         
            \coloredmatrix{0.07}{0.12}{0.03}{0.07}  \\
            History & 
                        \coloredmatrixdelta{0.42}{\down{.13}}{0.53}{\down{.14}}{0.51}{\down{.04}}{0.67}{\up{.11}} &
                        \coloredmatrixdelta{0.00}{\down{.01}}{0.00}{\same}{0.00}{\same}{0.00}{\same} &
                        \coloredmatrixdelta{0.46}{\down{.01}}{0.60}{\up{.10}}{0.48}{\down{.03}}{0.50}{\up{.04}} &
                        \coloredmatrixdelta{0.08}{\down{.01}}{0.08}{\down{.04}}{0.00}{\down{.03}}{0.05}{\down{.02}}  \\

            \bottomrule
        \end{tabularx}
    \caption{Feedback accuracy with the FM not given domain dynamics (Unk), given fully domain dynamics via text description (Full), given domain dynamics only for legal actions in the current state (Legal Action), and inferring dynamics from history context (History). For Robomimic, the last column is Delta action. For CliffWalking, the last column is Goal Advising.}
    \label{tab:dynamics}
    \vspace{-0.1in}
\end{table}

\paragraph{ICL and Thinking Guides respectively help bigger and reasoning models}
Tables~\ref{tab:all_icl_expert} and~\ref{tab:thinking_guides} show performance gains from ICL and Thinking Guides.
Smaller FMs generally do not benefit from either techniques and are occasionally confused by the additional complexities in the prompt.
For instance, Llama 3.1 8B degrades by $\sim$0.1 across feedback types in Door Key with ICL, and by as much as 0.21 in HierarchyCraft with Thinking Guides.
In contrast, for larger FMs, ICL gains tend to be consistently positive, with negative changes being very limited. 
For instance, in HeirarchyCraft, ICL improves action advising of R1 by 0.1 and preference of Llama 3.1 70B by 0.42.
Thinking Guides consistently improve the performance of reasoning FMs such as QwQ and R1, though by less than ICL.
On non-reasoning models, its gains are inconsistent across feedback types.
For VLMs, neither ICL nor Thinking Guides consistently improves performance, apart from ICL binary feedback for small VLMs in Qwen 2.5 VL 7B and 32B.

\paragraph{Access to environment dynamics hints can improve feedback accuracy.}
Table~\ref{tab:dynamics} shows FM performance given different hints about high-level domain transition dynamics.
Full descriptions of dynamics generally improve feedback quality, especially for small domains such as CliffWalking and Door Key.
For complex domains such as HierarchyCraft, long prompts that describe complex dynamics degrade the performance of small FMs, but consistently improve the performance of large FMs.
For instance, the 8B Llama feedback worsens across feedback types while the 70B model consistently improves, by as much as 0.24 for preference feedback. Similar trends hold when FMs induce dynamics from history context: large models benefit from context while small models do not.
A more significant improvement is observed in VLM feedback. Allowing access to history context allows even up to 22\% increase in action advising accuracy in Minigrid Doorkey and 44\% increase on binary feedback accuracy in Robomimic. In Door Key, feedback accuracy with history context and unknown dynamics shows similar performance with fully known dynamics, suggesting that the VLM is able to infer dynamics from the history context.

\paragraph{Analysis of wrong responses}
We identify main types of feedback errors as \textbf{format}, \textbf{insufficient knowledge}, \textbf{incorrect reasoning}, and \textbf{incorrect visual understanding} and show detailed examples in Appendix~\ref{app:response_examples}.
\textbf{Format} errors overwhelmingly occur with small FMs, which are are prone to output invalid JSON and invent answers not in the given options (e.g~\texttt{BOTH} during preference feedback).
\textbf{Insufficient knowledge} errors occur when FMs lack the world knowledge necessary to deduce the correct judgment.
This type of error occurs most in complex domains such as HierarchyCraft, where FMs express incorrect assumptions about the domain, and reason (correctly) towards the wrong conclusion.
A less frequent source of such errors is from powerful FMs that make extraneous assumptions about the domain.
For example, an action is preferable in Cliff Walking if it reduces the number of steps to the goal, yet R1 sometimes misclassifies optimal actions because R1 deems it important to maintain a long distance from the cliff.
Giving the FMs hints about domain dynamics reduces the frequency of these errors.
\textbf{Incorrect reasoning} errors occur when the FM demonstrate a correct understanding of domain dynamics, but incorrectly reasons towards a wrong conclusion.
This is common in domains with complex state spaces such as continuous sensors in Robomic, where FMs tend to correctly state domain dynamics but incorrectly calculate state transitions.
For instance, the FM may calculate the incorrect end effector location in Robomic after taking a delta action despite knowing the formula, or picking up unnecessary items in HierarchyCraft despite knowing the correct recipes. \textbf{Incorrect visual understanding} happens when the vision language model cannot correctly understand the image observation. For example, in Door Key, weaker VLMs frequently misinterpret the orientation of the agent and the location of walls.

\section{Related Work}

\paragraph{Learning from feedback}
Learning from feedback pre-dates LLMs.
\citet{rossReductionImitationLearning2011} directly uses expert demonstrations to learn policies.
\citet{ho2016generative} induce policies from offline expert behaviour, without interacting with experts during training.
\citet{torabi2018generative} induce policies from observations in expert trajectories, without access to annotated expert actions.
Instead of direct policy induction, expert behaviour can also be used to infer reward functions in inverse reinforcement learning~\citep{ng2000algorithms}, which can then be used to train policies.
Instead of expert behaviour, several recent works instead use expert preference to infer reward functions~\citep{ouyangTrainingLanguageModels2022a, lee2021pebble, metcalfSampleEfficientPreferencebasedReinforcement2024}. Vision language models have also been used to provide preference for RL~\citep{wang2024, rlsallmf}.
Compared to prior work, we evaluate LLM feedback without expert policies and without annotated demonstrations.
Extracting high quality automatic feedback significant reduces the amount of human labour required for training.

\paragraph{Learning from AI feedback}
Instead of extracting feedback from domain experts, recent work extract feedback from domain-agnostic pretrained foundation models.
\citet{Bai2022-constitutionalfeedback} outline a framework for alignment using AI feedback during pretraining and fine-tuning of LLMs.
\citet{Du2023-guidingmodels} use LLMs to generate intermediate goals for reinforcement learning.
\citet{xie2023text2reward} and~\citet{Wang2024-rl-vlm-ffeedback} generate reward functions respectively using LLMs and vision-language models.
\citet{colas2023AugmentingAutotelicAgents} use LLMs to retroactive relabel actions.
\citet{zhongPolicyImprovementUsing} use LLMs to identify desirable behaviours for imitation~\citep{zhongPolicyImprovementUsing}.
Several works use LLMs to generate preference data for exploration~\citep{Klissarov2023-motiffeedback} and policy search~\citep{Lee2023-rlaiffeedback,Liu2024-directdistillation}.
\citet{Sharma2024-criticalmodels} critically evaluate this last family of preference approaches, noting that gains largely result from combining superior teachers and weaker students. 
Our work is a unified, comprehensive study of AI feedback approaches, specifically evaluating the limitations of current state-of-the-art LLMs for providing planning feedback in grounded environments.
We consider the primary types AI feedback in prior works, and evaluate these techniques on a shared, diverse set of domains across model classes and \inferenceconditions.

\paragraph{LLMs for planning}
LLMs capture world knowledge helpful for planning~\citep{brown2020language,liuInteractiveRobotLearning2023,stamperEnhancingLLMBasedFeedback2024,wuSymbolLLMLeverageLanguage,deepseekai2025deepseekr1incentivizingreasoningcapability}.
\citet{wangVoyagerOpenEndedEmbodied2023} use LLMs as policy for learning Minecraft.
\citet{saycan2022arxiv} and~\citet{vempralaChatGPTRoboticsDesign2024} use LLMs to compose executable code using predefined APIs for robotic control.
These are but a small representative sample recent works that use LLMs as a policy.
LLMs can also be used to provide action priors~\citep{zhangHowCanLLM2024}, subgoals~\citep{Du2023-guidingmodels}, and reward functions~\citep{xie2023text2reward} for reinforcement learning.
These methods for using LLMs in planning are two specific types of feedback considered in our study.
The action advising feedback is identical to using LLM as policy.
The goal advising feedback is similar to using LLM for next-goal generation or value estimation.
Beyond action and goal advising, we evaluates how well LLMs can provide other types of feedback under diverse \inferenceconditions~and environment settings.

\section{Conclusion and future work}

We conduct a comprehensive study on evaluating language model feedback across feedback types, environment domains, and \inferenceconditions.
We find that LLM feedback quality improves with larger models and reasoning models, with large reasoning models providing diverse feedback with 80+\% accuracy for most domains.
These models also exhibit less class and positional bias.
Inference-time techniques such as in-context learning, chain-of-thought, and dynamics hints improve the performance of large reasoning models, however the complex prompt injections they require degrade performance of smaller or non-reasoning models. 
For domains such as continuous control, where it is challenging to provide succinct textual descriptions of state and actions, VLMs can provide higher quality feedback than LLMs.
Given the evidence of high-quality LLM feedback, we encourage future research in how to leverage such feedback to systematically improve planning sample efficiency and quality.
In addition, future study is required in how to improve LLM feedback in domains with complex dynamics and with continuous state and action spaces --- both common characteristics of challenging environments for which it is even more difficult to engineer rewards and label demonstrations.

\section*{Ethics Statement}
Though currently we only evaluate LLM feedback quality in grounded domains, the released feedback dataset can be used to train decision-making policies.
This feedback dataset can be misused (e.g.~training to reinforce incorrect feedback) to mislead policy training or create misleading feedback models, both of which may result in harmful model behaviour if deployed.
We consider this work as a first step towards understanding LLM feedback, such that we can quantify limitations and mitigate risks when using such feedback to train models with less human annotation and fewer labelled demonstrations.

\bibliography{conference}
\bibliographystyle{conference}

\appendix
\clearpage
\section{Egocentric Observation}
\label{app:egocentric}
\citet{labash2020perspective} show that egocentric observation ability as well as perspective-taking ability is important for policy learning~\citep{labash2020perspective}.
For LLMs specifically,~\citet{szot2023large} and~\citet{zhongPolicyImprovementUsing} show that egocentric verbalization improves the performance of LLM agents in embodied planning.
Table~\ref{tab:egocentric} shows that unlike prior results for planning, egocentric verbalization generally degrades feedback performance across feedback types and across most domains.
\begin{table}[h]
    \centering
    \begin{tabularx}{\linewidth}{lXXXX}
        \toprule
        Domain & Preference & Action Adv & Bin Feedback  &Delta Act\\
        \midrule        
            Cliff Walking & 
                        \coloredmatrixdelta{0.50}{\up{0.05}}{0.55}{\down{ 0.08}}{0.95}{\up{.04}}{1.00}{\same} &
                        \coloredmatrixdelta{0.44}{\down{0.14}}{0.49}{\down{ 0.14}}{1.00}{\same}{1.00}{\same} &
                        \coloredmatrixdelta{0.50}{\down{0.04}}{0.50}{\down{0.09}}{0.88}{\down{.12}}{0.89}{\down{.11}} &
                        N/A 
            \\
            \midrule
            Door Key & 
                        \coloredmatrixdelta{0.53}{\down{.08}}{0.72}{\down{.03}}{0.67}{\down{.12}}{0.70}{\down{.28}} &
                        \coloredmatrixdelta{0.52}{\up{.08}}{0.30}{\down{.13}}{0.95}{\up{.14}}{1.00}{\up{.01}} &
                        \coloredmatrixdelta{0.80}{\up{.02}}{0.72}{\down{.03}}{0.73}{\down{.19}}{0.76}{\down{.24}} &
                        N/A 
            \\
                        \midrule
            Four Rooms & 
                        \coloredmatrixdelta{0.37}{\down{.22}}{0.63}{\down{.03}}{0.72}{\up{.07}}{0.69}{\down{.06}} &
                        \coloredmatrixdelta{0.34}{\up{.02}}{0.33}{\down{.11}}{0.83}{\up{.22}}{0.83}{\up{.05}} &
                        \coloredmatrixdelta{0.60}{\up{.01}}{0.68}{\down{.01}}{0.70}{\up{.04}}{0.72}{\down{0.09}} &
                        N/A 
            \\
            
                        \midrule
            Robomimic & 
                        \coloredmatrixdelta{0.52}{\down{.03}}{0.53}{\down{.14}}{0.47}{\down{.08}}{0.45}{\down{.11}} &
                        \coloredmatrixdelta{0.00}{\down{.01}}{0.00}{\same}{0.00}{\same}{0.00}{\same} &
                        \coloredmatrixdelta{0.52}{\up{.05}}{0.52}{\up{.02}}{0.49}{\down{.03}}{0.52}{\up{.06}} &
                        \coloredmatrixdelta{0.15}{\up{.08}}{0.06}{\down{.06}}{0.06}{\up{.03}}{0.07}{\same} 
            \\
            \bottomrule
        \end{tabularx}
    \caption{Feedback accuracy with egocentric observations.}
    \label{tab:egocentric}
\end{table}

\clearpage
\section{Results under different distributions}
\label{app:otherpolicies}
In this subsection, we present similar results acquired with datasets collected from other sampling policies (random policy and half-expert policy).
The results under the half-expert dataset and the random dataset are shown in Table~\ref{tab:all_half} and Table~\ref{tab:all_random}, and show similar trends to results under the expert dataset.
We also observe random dataset results seem to have a larger variance since the dataset's state distribution often only covers a small part of the state space.
For example, in Cliff Walking, a random policy has 75\% of the chance to take any of the actions \texttt{RIGHT}, \texttt{LEFT} or \texttt{DOWN} which results in hitting the wall or falling off the cliff.
Consequently, under the random policy, the agent almost never reaches the goal and mostly wanders around its start point.
The image observation results on VLMs under the half-expert dataset and the random dataset are shown in Table~\ref{tab:all_image_random} and Table~\ref{tab:all_image_half}. The results, similar to LLM feedback, are consistent with results in the expert dataset, while the random dataset poses more challenges, as it contains unusual states. For example, the action advising accuracy for Qwen 2.5 VL 72B dropped to 8\%.

\begin{table}[!t]
    \centering
    \begin{tabularx}{\linewidth}{lXXXX}
        \toprule
        Domain & Preference & Action Adv & Bin Feedback & Delta Action\\
        \midrule
    Door Key &
            \coloredmatrix{0.56}{0.73}{0.96}{0.99}
                    &
            \coloredmatrix{0.24}{0.29}{0.80}{0.99}
                    &
            \coloredmatrix{0.72}{0.66}{0.89}{0.95}
                    & \notapplicable
                    \\
                    \midrule
    Four Rooms &
            \coloredmatrix{0.51}{0.59}{0.64}{0.67}
                    &
            \coloredmatrix{0.32}{0.33}{0.40}{0.54}
                    &
            \coloredmatrix{0.57}{0.61}{0.66}{0.75}
                    & \notapplicable
                    \\
                    \midrule
    Robomimic &
            \coloredmatrix{0.53}{0.59}{0.31}{0.12}
                    &
            \coloredmatrix{0.00}{0.00}{0.00}{0.00}
                    &
            \coloredmatrix{0.54}{0.59}{0.31}{0.24}
                    & 
            \coloredmatrix{0.07}{0.09}{0.00}{0.04}
                    \\
                    \midrule
    ALFWorld &
            \coloredmatrix{0.50}{0.62}{0.71}{0.77}
                    &
            \coloredmatrix{0.52}{0.52}{0.52}{0.52}
                    &
            \coloredmatrix{0.50}{0.62}{0.71}{0.77}
                    & \notapplicable
                    \\
                    \midrule
    HierachyCraft &
            \coloredmatrix{0.48}{0.76}{0.84}{0.86}
                    &
            \coloredmatrix{0.53}{0.64}{0.66}{0.71}
                    &
            \coloredmatrix{0.49}{0.59}{0.58}{0.60}
                    & \notapplicable 
                    \\
        \bottomrule

\end{tabularx}
    \caption{
    Feedback accuracy in random dataset for \colorbox{red!30}{Llama 3.1 8B}, \colorbox{blue!30}{Llama 3.1 70B}, \colorbox{green!30}{QwQ 32B}, and \colorbox{yellow!30}{R1}.
    }
    \label{tab:all_random}
\end{table}
\begin{table}[!t]
    \centering
    \begin{tabularx}{\linewidth}{lXXXX}
        \toprule
        Domain & Preference & Action Adv & Bin Feedback & Delta Action\\
        \midrule
    Door Key &
            \coloredmatrix{0.57}{0.71}{0.91}{0.98}
                    &
            \coloredmatrix{0.37}{0.43}{0.82}{0.99}
                    &
            \coloredmatrix{0.78}{0.71}{0.91}{0.99}
                    & \notapplicable
                    \\
                    \midrule
    Four Rooms &
            \coloredmatrix{0.47}{0.59}{0.73}{0.75}
                    &
            \coloredmatrix{0.32}{0.38}{0.56}{0.74}
                    &
            \coloredmatrix{0.59}{0.64}{0.71}{0.77}
                    & \notapplicable
                    \\
                    \midrule
    Robomimic &
            \coloredmatrix{0.60}{0.63}{0.51}{0.23}
                    &
            \coloredmatrix{0.00}{0.00}{0.00}{0.00}
                    &
            \coloredmatrix{0.48}{0.51}{0.46}{0.27}
                    &
            \coloredmatrix{0.06}{0.13}{0.04}{0.03}
                    \\
                    \midrule
    ALFWorld &
            \coloredmatrix{0.52}{0.71}{0.75}{0.85}
                    &
            \coloredmatrix{0.49}{0.64}{0.78}{0.77}
                    &
            \coloredmatrix{0.77}{0.80}{0.83}{0.83}
                    & \notapplicable
                    \\
                    \midrule
    HierachyCraft &
            \coloredmatrix{0.51}{0.81}{0.85}{0.89}
                    &
            \coloredmatrix{0.39}{0.60}{0.64}{0.77}
                    &
            \coloredmatrix{0.76}{0.82}{0.84}{0.88}
                    & \notapplicable 
                    \\
        \bottomrule
    \end{tabularx}
    \caption{
    Feedback accuracy in half-expert dataset for \colorbox{red!30}{Llama 3.1 8B}, \colorbox{blue!30}{Llama 3.1 70B}, \colorbox{green!30}{QwQ 32B}, and \colorbox{yellow!30}{R1}.
    }
    \label{tab:all_half}
\end{table}
\begin{table}[!t]
    \centering
    \begin{tabularx}{\linewidth}{lXXXX}
        \toprule
        Domain & Preference & Action Adv & Bin Feedback & Delta Action\\
        \midrule
    Door Key &
            \coloredmatrix{0.27}{0.56}{0.52}{0.46}
                    &
            \coloredmatrix{0.22}{0.48}{0.08}{0.28}
                    &
            \coloredmatrix{0.55}{0.72}{0.62}{0.64}
                    & \notapplicable
                    \\
                    \midrule
    Robomimic &
            \coloredmatrix{0.92}{0.76}{0.85}{0.57}
                    &
            \coloredmatrix{0.00}{0.00}{0.00}{0.00}
                    &
            \coloredmatrix{0.38}{0.17}{0.06}{0.07}
                    & 
            \coloredmatrix{0.02}{0.08}{0.07}{0.00}
                    \\
        \bottomrule
    \end{tabularx}
    \caption{
    Feedback accuracy in random dataset for \colorbox{red!30}{Qwen 2.5 VL 7B}, \colorbox{blue!30}{Qwen 2.5 VL 32B}, \colorbox{green!30}{Qwen 2.5 VL 72B}, and \colorbox{yellow!30}{o4-mini}.
    }
    \label{tab:all_image_random}
    \vspace{-0.1in}
\end{table}

\begin{table}[!t]
    \centering
    \begin{tabularx}{\linewidth}{lXXXX}
        \toprule
        Domain & Preference & Action Adv & Bin Feedback & Delta Action\\
        \midrule
    Door Key &
            \coloredmatrix{0.37}{0.53}{0.52}{0.40}
                    &
            \coloredmatrix{0.39}{0.24}{0.28}{0.31}
                    &
            \coloredmatrix{0.50}{0.65}{0.63}{0.69}
                    & \notapplicable
                    \\
                    \midrule
    Robomimic &
            \coloredmatrix{0.85}{0.81}{0.82}{0.72}
                    &
            \coloredmatrix{0.00}{0.00}{0.00}{0.00}
                    &
            \coloredmatrix{0.42}{0.24}{0.01}{0.03}
                    & 
            \coloredmatrix{0.01}{0.09}{0.06}{0.02}
                    \\
        \bottomrule
    \end{tabularx}
    \caption{
    Feedback accuracy in half-expert dataset for \colorbox{red!30}{Qwen 2.5 VL 7B}, \colorbox{blue!30}{Qwen 2.5 VL 32B}, \colorbox{green!30}{Qwen 2.5 VL 72B}, and \colorbox{yellow!30}{o4-mini}.
    }
    \label{tab:all_image_half}
    \vspace{-0.1in}
\end{table}

\clearpage
\section{Positive vs Negative Bias in Binary Feedback}
\label{app:f1scores}
In this subsection, we evaluate Feedback Model binary feedback performance using F1 score in order to evaluate model performance in the event of class imbalance.
This imbalance in all datasets, since each domain has an action space with more than two actions.
Consequently, for binary feedback, a larger portion of examples contain non-optimal actions.
Table~\ref{tab:f1_expert} shows binary feedback F1 scores across domains.
We see that non-reasoning models, particularly Llama 3.1 8b, often obtain significantly worse recall than precision across domains.
In other words, non-reasoning models are strongly biased towards identifying actions as non-optimal.

\begin{table}[t]
    \centering
    \begin{tabularx}{\linewidth}{lXXX}
        \toprule
        Domain & F1 &Precision & Recall\\
        \midrule
        Cliff Walking &
            \coloredmatrix{0.44}{0.88}{0.66}{0.93} &
            \coloredmatrix{1.00}{1.00}{1.00}{1.00} &
            \coloredmatrix{0.28}{0.33}{0.42}{0.79}
                    \\
                    \midrule
    Door Key &
            \coloredmatrix{0.12}{0.51}{0.83}{1.00} &
            \coloredmatrix{0.47}{0.46}{0.89}{0.98} &
            \coloredmatrix{0.07}{0.57}{0.90}{1.00}
                    \\
                    \midrule
    Four Rooms &
            \coloredmatrix{0.28}{0.56}{0.53}{0.71}&
            \coloredmatrix{0.37}{0.55}{0.49}{0.66} &
            \coloredmatrix{0.22}{0.56}{0.57}{0.76}
                    \\
                    \midrule
    Robomimic &
            \coloredmatrix{0.05}{0.13}{0.00}{0.00}&
            \coloredmatrix{0.52}{0.68}{0.00}{0.00} &
            \coloredmatrix{0.03}{0.07}{0.00}{0.00} \\
                    \midrule
    ALFWorld &
            \coloredmatrix{0.18}{0.37}{0.29}{0.25}&
            \coloredmatrix{0.66}{0.49}{0.63}{0.66} &
            \coloredmatrix{0.11}{0.29}{0.19}{0.15}
                    \\
                    \midrule
    HierachyCraft &
            \coloredmatrix{0.31}{0.47}{0.55}{0.63} &
            \coloredmatrix{0.34}{0.37}{0.51}{0.59} &
            \coloredmatrix{0.28}{0.65}{0.60}{0.67}
                    \\
        \bottomrule
    \end{tabularx}
    \caption{
    F1 scores in traverse dataset (Cliff Walking) or expert dataset (all other domains) for \colorbox{red!30}{Llama 3.1 8B}, \colorbox{blue!30}{Llama 3.1 70B}, \colorbox{green!30}{QwQ 32B}, and \colorbox{yellow!30}{R1}.
    }
    \label{tab:f1_expert}
\end{table}

\clearpage
\section{Positional bias in pairwise preferences}
\label{app:positinalbias}
In our dataset collection process for preference, we enumerate all possible combinations of legal actions at each state.
Because of the sheer size of the dataset, we randomly downsample 1000 samples for experiments.
We conduct an extra study to investigate biases in positional preferences by Feedback Model in providing preference feedback.
Specifically, for two options $A$ and $B$, where $A$ is the ground truth preferable option, we ensure that both $(A, B)$ and $(B, A)$ are in the evaluation dataset.
The results are shown in Table~\ref{tab:pair}.
We find that smaller models like Llama 3.1 8B significantly favour whichever option is presented first, consequently resulting in high accuracy when the correct option is first but low accuracy when the correction option is second.
This bias is significantly reduced for bigger models, which are largely agnostic to the position of the correct option.
In other words, smaller models exhibit stronger positional bias than larger models.

\begin{table}[!t]
    \centering
    \begin{tabularx}{\textwidth}{lXXXXX}
        \toprule
        Location of ground truth option & Llama 3.1 8B accuracy & Llama 3.1 70B accuracy\\
        \midrule
        First & 0.76 &0.23 \\
        \midrule
        Second &0.69 &0.64 \\
        \bottomrule
    \end{tabularx}
    \caption{Positional bias of Llama 3.1 models on ALFWorld preference feedback using sample snapshots from expert policy}
    \label{tab:pair}
\end{table}

\clearpage
\section{Verbalisation and Prompt Examples}
\label{app:prompt_examples}
In this section, we provide prompt examples for selected domains and types of feedback using chain-of-thought prompting. We also provide prompt examples for different experimental conditions.

\subsection{Standard Prompts}

In this section, we provide prompt examples corresponding to Table~\ref{tab:all_normal_expert}, with no specific inference enhancement applied.
One relevant experiment we ran for binary feedback specifically considers choices for how to describe optimal actions in the prompt.
Our early results showed that asking ``is this action good or bad'' lacks precision and resulted identification of desirable behaviour that are not optimal as ``good''.
Our final experiments instead asks the Feedback Model ``is action \texttt{ACTION} the best action you can take?'', which resulted in significantly fewer false positives.

\begin{tcolorbox}[colback=red!5!white,colframe=red!75!black,title=Binary Feedback in MiniGrid Door Key]
You need to solve Minigrid Doorkey.\\
Minigrid Doorkey involves crossing a gridworld, where you need to pick up the key, unlock the door and reach the goal. Do not hit the wall.\\
The state is described with an ASCII art. In the ASCII art, ``\#'' means wall. ``K'' means key. ``D'' mean door. ``G'' means goal. ``.'' means empty tile. ``\^{ }'' mean the agent who is facing upwards. ``V'' mean the agent who is facing downwards. ``\textless{}'' means the agent who is facing leftwards. ``\textgreater{}'' means the agent who is facing rightwards.\\
The possible actions are: \\
1. \textbf{TURN LEFT}, changing the direction to left, for example, the agent ``\^{ }'' will become ``\textless{}''.\\
2. \textbf{TURN RIGHT}, changing the direction to right, for example, the agent ``\^{ }'' will become ``\textgreater{}''.\\
3. \textbf{MOVE FORWARD}, moving forwards for one tile\\
4. \textbf{PICK UP THE KEY}, picking up the key, this is only possible when you are facing the tile with key on it\\
5. \textbf{UNLOCK THE DOOR}, this action is only possible when you have the key.\\
\hspace*{1em}You will be asked to judge an action whether it is the optimal action, given a state.\\
You are in the following state represented in ASCII art:\\
\begin{verbatim}
#####
#.#.#
#>D.#
#.#G#
#####
\end{verbatim}
The door is locked. You have the key.\\
Is action \textbf{UNLOCK THE DOOR} the best action you can take? Please think step by step.\\
Only give the answer in a new line in JSON format:\\
\texttt{\{"reasoning": <REASONING>, "feedback": <FEEDBACK>\}}\\
Where \texttt{<FEEDBACK>} is one of ``YES'' or ``NO'', \texttt{<REASONING>} is a string of your thinking steps.

\end{tcolorbox}

\begin{tcolorbox}[colback=red!5!white,colframe=red!75!black,title=Preference in ALF World]
\small
You received instruction: Put a chilled potato in the microwave...
Your observation is: You open the fridge 1. The fridge 1 is open. In it, you see a egg 1, a lettuce 2, and a pan 1.
    
Below is your history:

You received instruction: Put a chilled potato in the microwave.
Your observation is: You are in the middle of a room. Looking quickly around you, you see a cabinet 6, a cabinet 5, a cabinet 4, a cabinet 3, a cabinet 2, a cabinet 1, a coffeemachine 1, a countertop 3, a countertop 2, a countertop 1, a drawer 3, a drawer 2, a drawer 1, a fridge 1, a garbagecan 1, a microwave 1, a shelf 3, a shelf 2, a shelf 1, a sinkbasin 1, a stoveburner 4, a stoveburner 3, a stoveburner 2, a stoveburner 1, and a toaster 1.
.
    You took action: look.

    You received instruction: Put a chilled potato in the microwave.
    Your observation is: You are in the middle of a room. Looking quickly around you, you see nothing..
    You took action: go to sinkbasin 1.

    You received instruction: Put a chilled potato in the microwave.
    Your observation is: You arrive at loc 29. On the sinkbasin 1, you see a butterknife 3, a cup 1, a egg 2, a knife 1, a spatula 2, and a tomato 1..
    You took action: go to microwave 1.

    You received instruction: Put a chilled potato in the microwave.
    Your observation is: You arrive at loc 17. The microwave 1 is closed..
    You took action: open microwave 1.

    You received instruction: Put a chilled potato in the microwave.
    Your observation is: You open the microwave 1. The microwave 1 is open. In it, you see a apple 1, and a mug 1..
    You took action: close microwave 1.

    You received instruction: Put a chilled potato in the microwave.
    Your observation is: You close the microwave 1..
    You took action: go to garbagecan 1.

    You received instruction: Put a chilled potato in the microwave.
    Your observation is: You arrive at loc 31. On the garbagecan 1, you see a apple 2, and a glassbottle 2..
    You took action: go to fridge 1.

    You received instruction: Put a chilled potato in the microwave.
    Your observation is: You arrive at loc 30. The fridge 1 is closed.
    You took action: open fridge 1.

You can take the following actions:

close fridge 1

examine fridge 1

go to cabinet 1

go to cabinet 2

go to cabinet 3

go to cabinet 4

go to cabinet 5

go to cabinet 6

go to coffeemachine 1

go to countertop 1

go to countertop 2

go to countertop 3

go to drawer 1

go to drawer 2

go to drawer 3

go to garbagecan 1

go to microwave 1

go to shelf 1

go to shelf 2

go to shelf 3

go to sinkbasin 1

$<$Some possible action omitted here for ease of read$>$

Given go to countertop 2 or close fridge 1, which action is better? Please think step by step.
Only give the answer in a new line in JSON format:
\{"reasoning": $<$REASONING$>$, "preference": $<$PREFERENCE$>$\}
Where $<$PREFERENCE$>$ is one of "FIRST" or "SECOND", $<$REASONING$>$ is a string of your thinking steps.

\end{tcolorbox}

\begin{tcolorbox}[colback=red!5!white,colframe=red!75!black,title=Action Advising in Hierachy Craft]
You are playing MineCraft and you want to craft diamond.

You are in forest.

Items in your current zone are:

crafting table: 1

furnace: 1

You have the following items in your inventory:

wood plank: 3

wood pickaxe: 7

stone pickaxe: 12

iron ore: 3

You can take these actions:

pickup crafting table

pickup furnace

smelt iron ore with wood plank

craft stick

search for dirt at forest

search for wood at forest

search for cobblestone with stone pickaxe at forest

search for cobblestone with wood pickaxe at forest

move to swamp from forest

move to meadow from forest

move to underground from forest with stone pickaxe

Which action do you choose? Please think step by step.

Only give the answer in a new line in JSON format:
\{"reasoning": $<$REASONING$>$, "action": $<$ACTION$>$\}
Where $<$ACTION$>$ is one of possible actions, $<$REASONING$>$ is a string of your thinking steps.

\end{tcolorbox}

\begin{tcolorbox}[colback=red!5!white,colframe=red!75!black,title=Delta Action in Robomimic]
\small

You are a robot engineer. You need to control a robot arm to lift an object. You need to observe the robot's current state and choose which action to do next in order to lift the object. You can move the robot in the X, Y, or Z direction, as well as open/close its gripper. The robot's state is given in the following format:
\begin{verbatim}
    Gripper position: [x y z ]
    Object position: [x y z]
    Object relative distance to gripper: [dx dy dz]
    Gripper finger state: open or closed
\end{verbatim}
You can choose actions that move the gripper in the X, Y, Z direction, with additional control of the gripper. The format of the action is:
\begin{verbatim}
[resulting x, resulting y, resulting z, ignore, ignore, ignore, resulting gripper state]
\end{verbatim}
Note on the gripper: Set grip to 1 to open the gripper and -1 to close the gripper. The robot's gripper should be closed if it is beginning to grasp the object, or when it is holding the object. When it is approaching the object, the gripper is open. If the robot gripper needs to be closed, you should continue to close the gripper, even if it is closed. Similarly, if the robot gripper needs to be open, you should continue to open the gripper, even if it is already open.\\

You current observation is:

Gripper position: [-0.01261643  0.02076922  0.82920151]
\\
Object position: [-0.02050141  0.02279622  0.82019789]
\\
Object relative position to gripper: [ 0.00788497 -0.002027    0.00900362]
\\
Gripper finger position: [ 0.03368153 -0.03366622]
\\
Gripper status: +1

Given the following actions, which action is better? Please think step by step.

Index 0: [ 0.25368214  0.02193805 -0.23200776  0.00478249  0.25400147  0.01591277  1.        ]
\\
Index 1: [-0.30631787  0.02193805 -0.23200776  0.00478249  0.25400147  0.01591277  1.        ]
\\
Index 2: [-0.02631785  0.30193806 -0.23200776  0.00478249  0.25400147  0.01591277  1.        ]
\\
Index 3: [-0.02631785 -0.25806195 -0.23200776  0.00478249  0.25400147  0.01591277  1.        ]
\\
Index 4: [-0.02631785  0.02193805  0.04799224  0.00478249  0.25400147  0.01591277  1.        ]
\\
Index 5: [-0.02631785  0.02193805 -0.5120078   0.00478249  0.25400147  0.01591277  1.        ]
\\
Index 12: [-0.02631785  0.02193805 -0.23200776  0.00478249  0.25400147  0.01591277  1.        ]
\\
Index 13: [-0.02631785  0.02193805 -0.23200776  0.00478249  0.25400147  0.01591277 -1.        ]

You should consider the position of the robot gripper and object, and how they are related to each other. For example, if the robot gripper is on the left of the object, you should consider moving the robot gripper to the right.\\
Only give the answer in a new line in JSON format:\\
\texttt{\{"reasoning": <REASONING>, "index": <INDEX>\}}\\
Where \texttt{<INDEX>} is one of action choice index, \texttt{<REASONING>} is a string of your thinking steps.
\end{tcolorbox}
\begin{tcolorbox}[colback=red!5!white,colframe=red!75!black,title=Goal Advising in Cliff Walking]
You are an agent trying to move to position (3, 11).\\
You can take 4 possible actions:
\begin{verbatim}
  - going UP
  - going DOWN
  - going LEFT
  - going RIGHT
\end{verbatim}

The grid world is 4x12 with zero-indexed positions. There are holes on the grid that you do not know. Do not exceed the boundaries of the world, and do not fall into the holes. You will be asked to predict the best next state given a state.\\

You are in position (3, 0). What coordinates should you move to in the next step? Please think step by step.\\
Only give the answer in a new line in JSON format:\\
\texttt{\{"reasoning": <REASONING>, "row": <ROW>, "column": <COLUMN>\}}\\
Where \texttt{<ROW>} is the number of the row you should move, and \texttt{<COLUMN>} is the number of column you should move, \texttt{<REASONING>} is a string of your thinking steps.
\end{tcolorbox}

\clearpage
\subsection{In Context Learning Prompt Examples}
In this section, we provide prompt examples for our ICL experiments. We insert these examples after the description of the task, right before providing the current observation and querying for feedback.

\begin{tcolorbox}[colback=red!5!white,colframe=red!75!black,title=Cliff Walking Goal Advising ICL Example]
Below are examples about reasoning over the best next state.\\

\textbf{Question:} You are in position (3, 0). What coordinates should you move to in the next step?\\
\textbf{Answer:} \texttt{\{"reasoning": "There are 4 neighbouring coordinates: (2, 0), (4, 0), (3, -1), (3, 1). If I move to (2, 0), it is a safe place. If I move to (4, 0), I will hit the wall as I am in a zero-indexed 4x12 grid and (4, 0) is outside of boundary. If I move to (3, -1), I will similarly hit the wall as (3, -1) is outside of boundary. If I move to (3, 1) and since there are holes in (3, 1) ... (3, 10), I will fall into a hole. Therefore moving to (2, 0) is the only safe option.","row": "2", "column": "0"\}}\\

\textbf{Question:} You are in position (2, 11). What coordinates should you move to in the next step?\\
\textbf{Answer:} \texttt{\{"reasoning": "There are 4 neighbouring coordinates: (1, 11), (3, 11), (2, 10) and (2, 12). If I move to (1, 11) and then I will need 2 steps as (1, 11) -{}> (2, 11) -{}> (3, 11) to reach the goal (3, 11). If I move to (3, 11) I will reach the goal. If I move to (2, 10) and then I need 3 steps as (2, 10) -{}> (3, 10) -{}> (3, 11) to reach the goal. If I move to (2, 12) I will hit the wall as it is outside of boundary. Therefore (3, 11) is the best option.","row": "3", "column": "11"\}}\\

\textbf{Question:} You are in position (1, 5). What coordinates should you move to in the next step?\\
\textbf{Answer:} \texttt{\{"reasoning": "There are 4 possible action: (0, 5), (2, 5), (1, 6), (1, 4). If I move to (0, 5) and I will need 9 steps to reach goal: (0, 5) -{}> (0, 6) -{}> (0, 7) -{}> (0, 8) -{}> (0, 9) -{}> (0, 10) -{}> (0, 11) -{}> (1, 11) -{}> (2, 11) -{}> (3, 11). If I move to (2, 5) and I will need 7 steps to reach goal: (2, 5) -{}> (2, 6) -{}> (2, 7) -{}> (2, 8) -{}> (2, 9) -{}> (2, 10) -{}> (2, 11) -{}> (3, 11). If I move to (1, 6) and I will need 7 steps to reach goal: (1, 6) -{}> (1, 7) -{}> (1, 8) -{}> (1, 9) -{}> (1, 10) -{}> (1, 11) -{}> (2, 11) -{}> (3, 11). If I move to (1, 4) and I will need 9 steps to reach goal: (1, 4) -{}> (1, 5) -{}> (1, 6) -{}> (1, 7) -{}> (1, 8) -{}> (1, 9) -{}> (1, 10) -{}> (1, 11) -{}> (2, 11) -{}> (3, 11). Therefore, (1, 6) and (2, 5) are the best next coordinates as they give shortest path to goal. I just need one good coordinates so I will pick one from (1, 6) and (2, 5)","row": "2", "column": "5"\}}\\
\end{tcolorbox}

\begin{tcolorbox}[colback=red!5!white,colframe=red!75!black,title=Door Key Binary Feedback ICL]
\textbf{Answer:} \texttt{\{"reasoning": "The agent is facing a wall, so moving forward does nothing and fails to progress toward picking up the key. A turn to face the key is a better action.", "feedback": "NO"\}}\\

\textbf{Question:} You are in the following state represented in ascii art: 
\begin{verbatim}
##### 
#.D.# 
#^#.# 
#.#G# 
#####
\end{verbatim}

The door is locked. You have the key.\\
Is action MOVE FORWARD the best action you can take? Please think step by step.\\
Only give the answer in a new line in JSON format:\\
\texttt{\{"reasoning": <REASONING>, "feedback": <FEEDBACK>\}}\\
Where \texttt{<FEEDBACK>} is one of ``YES'' or ``NO'', \texttt{<REASONING>} is a string of your thinking steps.\\

\textbf{Answer:} \texttt{\{"reasoning": "You already have the key, so the immediate goal is to move adjacent to the locked door and unlock it. The tile directly ahead is free, and moving forward places you in position to turn and unlock the door. Therefore, moving forward is the best action.", "feedback": "YES"\}}\\

\textbf{Question:} You are in the following state represented in ascii art: 
\begin{verbatim}
##### 
#.#.# 
#.>.# 
#.#G# 
##### 
\end{verbatim}
The door is unlocked. You have the key.\\
Is action TURN LEFT the best action you can take? Please think step by step.\\
Only give the answer in a new line in JSON format:\\
\texttt{\{"reasoning": <REASONING>, "feedback": <FEEDBACK>\}}\\
Where \texttt{<FEEDBACK>} is one of ``YES'' or ``NO'', \texttt{<REASONING>} is a string of your thinking steps.\\

\textbf{Answer:} \texttt{\{"reasoning": "The agent is facing right and needs to move toward the goal on its right and then turn down. Turning left would make the agent face upwards into a wall. Moving forward is the better move.", "feedback": "NO"\}}\\
\end{tcolorbox}

\begin{tcolorbox}[colback=red!5!white,colframe=red!75!black,title=Action Advising ICL Four Rooms]

Below are examples about reasoning over the best action.\\

\textbf{Question:}\\
You are in the following state represented in ascii art:\\
\begin{verbatim}
###################
#........#........#
#........#........#
#........#........#
#........#........#
#........#........#
#........#........#
#........#........#
#.................#
######.##########.#
#........#....G.<.#
#........#........#
#........#........#
#.................#
#........#........#
#........#........#
#........#........#
#........#........#
###################
\end{verbatim}
Which action do you choose? Please think step by step.\\
Only give the answer in a new line in JSON format:\\
\texttt{\{"reasoning": <REASONING>, "action": <ACTION>\}}\\
Where \texttt{<ACTION>} is one of ``TURN LEFT'', ``TURN RIGHT'' or ``MOVE FORWARD'', \texttt{<REASONING>} is a string of your thinking steps.\\

\textbf{Answer:} \texttt{\{"reasoning": "The agent is already facing left and the goal is two tiles to its left, so moving forward brings it closer to the goal.", "action": "MOVE FORWARD"\}}\\

\end{tcolorbox}

\begin{tcolorbox}[colback=red!5!white,colframe=red!75!black,title=HierachyCraft ICL in Preference]
\textbf{Question:}\\
You are playing MineCraft and you want to craft diamond. You will be asked to compare two actions given a state and provide preference.\\

You are in forest.\\
Items in your current zone are:
\begin{itemize}
  \item crafting table: 1
\end{itemize}

You have the following items in your inventory:
\begin{itemize}
  \item wood plank: 4
  \item stick: 8
\end{itemize}

You can take these actions:
\begin{itemize}
  \item pickup crafting table
  \item craft stick
  \item craft crafting table
  \item craft wood pickaxe
  \item craft wood axe
  \item craft wood shovel
  \item craft wood sword
  \item search for dirt at forest
  \item search for wood at forest
  \item move to swamp from forest
  \item move to meadow from forest
\end{itemize}

Given \texttt{craft crafting table} or \texttt{craft wood pickaxe}, which action is better? Please think step by step.\\
Only give the answer in a new line in JSON format:\\
\texttt{\{"reasoning": <REASONING>, "preference": <PREFERENCE>\}}\\
Where \texttt{<PREFERENCE>} is one of ``FIRST'' or ``SECOND'', \texttt{<REASONING>} is a string of your thinking steps.\\

\textbf{Answer:} \texttt{\{"reasoning": "We already have access to a crafting table in the forest, so crafting another is not immediately useful. We need a wood pickaxe to gather stone and progress toward diamonds.", "preference": "SECOND"\}}\\

\end{tcolorbox}

\begin{tcolorbox}[colback=red!5!white,colframe=red!75!black,title=ALF World ICL Binary Feedback Example]
\small
\textbf{Question:}\\
You received instruction: Place a chilled tomato in a microwave.\\
Your observation is: You arrive at loc 3. On the shelf 2, you see a creditcard 2.\\
Your inventory: tomato 1.
Below is your history:
\begin{verbatim}
You received instruction: Place a chilled tomato in a microwave.
Your observation is: You are in the middle of a room. Looking quickly around 
you, you see a cabinet 6, a cabinet 5, a cabinet 4, a cabinet 3, a cabinet 2, 
a cabinet 1, a coffeemachine 1, a countertop 3, a countertop 2, a countertop 1,
a drawer 3, a drawer 2, a drawer 1, a fridge 1, a garbagecan 1, a microwave 1,
a shelf 3, a shelf 2, a shelf 1, a sinkbasin 1, a stoveburner 4, a stoveburner 3,
a stoveburner 2, a stoveburner 1, and a toaster 1.
Your inventory: empty.
You took action: go to cabinet 1.

You received instruction: Place a chilled tomato in a microwave.
Your observation is: You arrive at loc 13. On the cabinet 1, you see a bowl 1.
Your inventory: empty.
You took action: go to countertop 1.

<More History Steps Ommited for Ease of Reading>
\end{verbatim}

You can take the following actions:\\
\texttt{examine shelf 2}\\
\texttt{examine tomato 1}\\
\texttt{go to cabinet 1}\\
\texttt{go to cabinet 2}\\
\texttt{go to cabinet 3}\\
\texttt{go to cabinet 4}\\
\texttt{go to cabinet 5}\\
\texttt{go to cabinet 6}\\
\texttt{go to coffeemachine 1}\\
\texttt{go to countertop 1}\\
\texttt{go to countertop 2}\\
\texttt{go to countertop 3}\\
\texttt{go to drawer 1}\\
\texttt{go to drawer 2}\\
\texttt{go to drawer 3}\\
\texttt{go to fridge 1}\\
\texttt{go to garbagecan 1}\\
\texttt{go to microwave 1}\\
\texttt{go to shelf 1}\\
\texttt{go to shelf 3}\\
\texttt{go to sinkbasin 1}\\
\texttt{go to stoveburner 1}\\
\texttt{go to stoveburner 2}\\
\texttt{go to stoveburner 3}\\
\texttt{go to stoveburner 4}\\
\texttt{go to toaster 1}\\
\texttt{inventory}\\
\texttt{look}\\

Is action \texttt{go to cabinet 3} the best action you can take? Please think step by step.\\
Only give the answer in a new line in JSON format:\\
\texttt{\{"reasoning": <REASONING>, "feedback": <FEEDBACK>\}}\\
Where \texttt{<FEEDBACK>} is one of ``YES'' or ``NO'', \texttt{<REASONING>} is a string of your thinking steps.\\

\textbf{Answer:} \texttt{\{"reasoning": "We already have the chilled tomato in our inventory and our goal is to place it in the microwave. The next logical step is to go back to the microwave and open it, then put the tomato inside. Going to cabinet 3 does not help achieve this goal.", "feedback": "NO"\}}\\
\end{tcolorbox}

\clearpage
\subsection{Thinking Guides Prompt Examples}

In this subsection, we show the example prompts of thinking guides. We insert these thinking guides inside standard prompts, right after we describe the domain and before we provide the current observation and query for feedback.

\begin{tcolorbox}[colback=red!5!white,colframe=red!75!black,title=Thinking Guides for Door Key and Four Rooms]
You should think about these questions.\\
What is your current goal?\\
Based on the rules you know, what subgoals do you need to achieve?\\
Based on what you see and what you have, what subgoals have you achieved?\\
If you have not achieved all of the subgoals, based on the rules and doable actions, can you achieve any subgoals now?\\
If yes, which subgoal do you want to achieve next?\\
If no, what do you need or where should you go next?
\end{tcolorbox}

\begin{tcolorbox}[colback=red!5!white,colframe=red!75!black,title=Thinking Guides for HierachyCraft]
You should think about these questions. \\
You need to get a diamond. \\
Based on the rules you know, what subgoals do you need to achieve? \\
Based on what you see and what you have, what subgoals have you achieved? \\
If you have not achieved all of the subgoals, based on the rules and doable actions, can you achieve any subgoals now? \\
If yes, which subgoal do you want to achieve next? \\
If no, what do you need or where should you go next? 
\end{tcolorbox}

\begin{tcolorbox}[colback=red!5!white,colframe=red!75!black,title=Thinking Guides for Cliff Walking]
You should think about these questions. \\
What is your current goal? \\
Based on the rules you know, which coordinates do you need to reach the goal? \\
Based on what you see and what you known, can you reach any of the coordinates now? \\
If no, comparing the possible actions you can take, which one leads you closer to the goal? 
\end{tcolorbox}

\begin{tcolorbox}[colback=red!5!white,colframe=red!75!black,title=Thinking Guides for ALF World]
You should think about these questions. \\
What is your current goal? \\
Based on the rules you know, what subgoals do you need to achieve? \\
Based on what you see and what you have, what subgoals have you achieved? \\
If you have not achieved all of the subgoals, based on the rules and doable actions, can you achieve any subgoals now? \\
If yes, which subgoal do you want to achieve next? \\
If no, what do you need or where should you go next? \\
\end{tcolorbox}

\clearpage
\subsection{Dynamics Hint Prompt Examples}
In this subsection, we show prompt examples of experiments with/without dynamics hints. For unknown dynamics, we alter the task description at the beginning of the prompt and remove anything related to dynamics.
For fully known dynamics, we append the rules of the game right after we describe the task before we provide states and query for feedback.
For history context, we append the history steps at the end of the prompt, right after we provide the current observation and query for feedback.

\begin{tcolorbox}[colback=red!5!white,colframe=red!75!black,title=Cliff Walking Unknown Dynamics]
You are an agent trying to move to position (3, 11). \\
You can take 4 possible actions: 
\begin{verbatim}
- going UP 
- going DOWN
- going LEFT
- going RIGHT 
\end{verbatim}
The grid world is 4x12 with zero-indexed positions. There are holes on the grid that you do not know. Do not exceed the boundaries of the world, and do not fall into the holes.
\end{tcolorbox}

\begin{tcolorbox}[colback=red!5!white,colframe=red!75!black,title=Cliff Walking History Context Example]
Here is the history of your actions:

Step 1: You were at position (3, 0), and you took action UP, then you reached (2, 0). 

Step 2: You were at position (2, 0), and you took action DOWN, then you reached (3, 0). 

Step 3: You were at position (3, 0), and you took action RIGHT, then you reached (3, 0). You fell into the holes.

\end{tcolorbox}

\begin{tcolorbox}[colback=red!5!white,colframe=red!75!black,title=Door Key Unknown Dynamics Example]
You need to solve Minigrid Doorkey.
Minigrid Doorkey involves crossing a gridworld, where you need to pick up the key, unlock the door and reach the goal. Do not hit the wall.
The state is described with an ASCII art. In the ASCII art, ``\#'' means wall. ``K'' means key. ``D'' means door. ``G'' means goal. ``.'' means empty tile. ``\^{ }'' means the agent who is facing upwards. ``V'' means the agent who is facing downwards. ``\textless{}'' means the agent who is facing leftwards. ``\textgreater{}'' means the agent who is facing rightwards.
The possible actions are:
\begin{verbatim}
    1. TURN LEFT
    2. TURN RIGHT
    3. MOVE FORWARD
    4. PICK UP THE KEY
    5. UNLOCK THE DOOR
\end{verbatim}
    
\end{tcolorbox}

\begin{tcolorbox}[colback=red!5!white,colframe=red!75!black,title=Door Key History Context Example]
Here is the history of your actions:\\
\textbf{Step 1:} You see:
\begin{verbatim}
#####
#K#.#
#.D.#
#<#G#
#####
\end{verbatim}
You don't have the key and the door is locked. You took action \texttt{TURN RIGHT}.\\
\textbf{Step 2:} You see:
\begin{verbatim}
#####
#K#.#
#.D.#
#^#G#
#####
\end{verbatim}
You don't have the key and the door is locked. You took action \texttt{MOVE FORWARD}.\\
\textbf{Step 3:} You see:
\begin{verbatim}
#####
#K#.#
#^D.#
#.#G#
#####
\end{verbatim}
You don't have the key. The door is locked. You took action \texttt{PICK UP THE KEY}.\\
\textbf{Step 4:} You see:
\begin{verbatim}
#####
#.#.#
#^D.#
#.#G#
#####
\end{verbatim}
You have the key. The door is locked. You took action \texttt{TURN RIGHT}.\\
\end{tcolorbox}

\begin{tcolorbox}[colback=red!5!white,colframe=red!75!black,title=HierachyCraft Full Dynamics Example]
Below are the rules for the minecraft.
\begin{verbatim}
You need a crafting table to craft something.
You need a furnace to smelt something.
You can pick up a crafting table or a furnace and take them around. 
You need to place one in your current area if you want to use them.
You can get 3 iron ingot after smelting with 3 iron ore, 2 wood plank.
You can get 3 gold ingot after smelting with 3 gold ore, 2 wood plank.
You can get 8 iron ingot after smelting with 8 iron ore, 1 coal.
You can get 8 gold ingot after smelting with 8 gold ore, 1 coal.
You can craft 4 wood plank with 1 wood.
You can craft 4 stick with 2 wood plank.
You can craft 1 furnace with 8 cobblestone.
You can craft 3 paper with 3 reeds.
You can craft 1 book with 3 paper, 1 leather.
You can craft 1 enchanting table with 1 book, 4 obsidian, 2 diamond.
You can craft 1 clock with 4 gold ingot, 1 redstone.
You can craft 1 flint with 10 gravel.
You can craft 4 flint and steel with 1 iron ingot, 1 flint.
You can craft 2 blaze powder with 1 blaze rod.
You can craft 1 ender eye with 1 blaze powder, 1 ender pearl.
You can craft 10 wood pickaxe with 3 wood plank, 2 stick.
You can craft 10 wood axe with 3 wood plank, 2 stick.
You can craft 10 wood shovel with 1 wood plank, 2 stick.
You can craft 10 wood sword with 2 wood plank, 1 stick.
You can craft 20 stone pickaxe with 3 cobblestone, 2 stick.
You can craft 20 stone axe with 3 cobblestone, 2 stick.
You can craft 20 stone shovel with 1 cobblestone, 2 stick.
You can craft 20 stone sword with 2 cobblestone, 1 stick.
You can craft 40 iron pickaxe with 3 iron ingot, 2 stick.
You can craft 40 iron axe with 3 iron ingot, 2 stick.
You can craft 40 iron shovel with 1 iron ingot, 2 stick.
You can craft 40 iron sword with 2 iron ingot, 1 stick.
You can craft 10 gold pickaxe with 3 gold ingot, 2 stick.
You can craft 10 gold axe with 3 gold ingot, 2 stick.
You can craft 10 gold shovel with 1 gold ingot, 2 stick.
You can craft 10 gold sword with 2 gold ingot, 1 stick.
You can craft 100 diamond pickaxe with 3 diamond, 2 stick.
You can craft 100 diamond axe with 3 diamond, 2 stick.
You can craft 100 diamond shovel with 1 diamond, 2 stick.
You can craft 100 diamond sword with 2 diamond, 1 stick.
You can collect 2 dirt at forest.
You can collect 2 dirt at swamp.
You can collect 2 dirt at meadow.
You need to have at least 1 diamond shovel to collect 16 dirt at forest. 
1 diamond shovel will be consumed.
<The rest are omitted>

\end{verbatim}
\end{tcolorbox}

\clearpage
\subsection{Egocentric Prompt Examples}
In MiniGrids Egocentric Observation experiments, the ASCII arts are rotated so that the agent is always facing forward.
In Robomimic experiments, we verbalise the observation with relative position to the object.

\begin{tcolorbox}[colback=red!5!white,colframe=red!75!black,title=Door Key with Egocentric Observation]
You need to solve Minigrid Doorkey.\\
Minigrid Doorkey involves crossing a gridworld, where you need to pick up the key, unlock the door and reach the goal. Do not hit the wall.\\
The state is described with an ASCII art in first-person perspective. In the ASCII art, ``\#'' means wall. ``K'' means key. ``D'' means door. ``G'' means goal. ``.'' means empty tile.\\
``\^{ }'' is where you are. You are always facing forward.\\
The possible actions are:
\begin{enumerate}
  \item TURN LEFT.
  \item TURN RIGHT.
  \item MOVE FORWARD, moving forwards for one tile
  \item PICK UP THE KEY, picking up the key, this is only possible when you are facing the tile with key on it.
  \item UNLOCK THE DOOR, this action is only possible when you have the key.
\end{enumerate}
You will be asked to predict the best action given a state.\\
You are in the following state represented in ASCII art:
\begin{verbatim}
#####
#..G#
##D##
#.^.#
#####
\end{verbatim}
The door is locked. You have the key.\\
Which action do you choose? Please think step by step.\\
Only give the answer in a new line in JSON format:\\
\texttt{\{"reasoning": <REASONING>, "action": <ACTION>\}}\\
Where \texttt{<ACTION>} is one of ``TURN LEFT'', ``TURN RIGHT'', ``MOVE FORWARD'', ``PICK UP THE KEY'' or ``UNLOCK THE DOOR'', \texttt{<REASONING>} is a string of your thinking steps.
\end{tcolorbox}

\begin{tcolorbox}[colback=red!5!white,colframe=red!75!black,title=Door Key with Egocentric Observation]
You are a robot arm trying to lift an object. You can move your gripper in the X, Y, or Z direction, as well as open/close your gripper. Each step, you see:
\begin{itemize}
  \item Relative distance from your gripper to object in x direction
  \item Relative distance from your gripper to object in y direction
  \item Relative distance from your gripper in z direction
  \item Gripper finger state: open or closed
\end{itemize}

You can choose actions that move the gripper in the X, Y, Z direction, with additional control of the gripper.\\

\textbf{Note on the gripper:} You should close it when beginning to grasp the object, or when you are holding the object. When you are approaching the object, the gripper should be open. If the gripper needs to be closed, you should continue to close the gripper, even if it is closed. Similarly, if the gripper needs to be open, you should continue to open the gripper, even if it is already open.\\

\textbf{You current observation is:}
\begin{itemize}
  \item The object is 0.002046926351863232 to the right of your gripper.
  \item The object is 0.024008545831402408 in front of your gripper.
  \item The object is 0.8217841891593334 below your gripper.
\end{itemize}

Given the following action:
[-0.2087612  -0.02484818 -0.16213629  0.01804361 -0.04799016  0.00671699  0.9998945 ]

Here is a list of corrective actions that you can take. Each action will modify the current gripper state by the action amount. Which action do you choose? Please think step by step.\\

\begin{itemize}
  \item Action 0: Move gripper forward by 0.28
  \item Action 1: Move gripper backward by 0.28
  \item Action 2: Move gripper right by 0.28
  \item Action 3: Move gripper left by 0.28
  \item Action 4: Move gripper vertically upward by 0.28
  \item Action 5: Move gripper vertically downward by 0.28
  \item Action 6: Open gripper
  \item Action 7: Close gripper
\end{itemize}

You should consider the position of the robot gripper and object, and how they are related to each other. For example, if the robot gripper is on the left of the object, you should consider moving the robot gripper to the right.\\

Only give the answer in a new line in JSON format:\\
\texttt{\{"reasoning": <REASONING>, "index": <INDEX>\}}\\
Where \texttt{<INDEX>} is one of action choice index, \texttt{<REASONING>} is a string of your thinking steps.\end{tcolorbox}

\clearpage
\subsection{VLM Feedback Prompt Examples}

\begin{tcolorbox}[colback=red!5!white,colframe=red!75!black,title=Door Key with Image Observation]
You need to solve Minigrid Doorkey. Minigrid Doorkey involves crossing a gridworld, where you need to pick up the key, unlock the door and reach the goal. Do not hit the wall. The state is depicted in an image. 

This is you facing upwards. \includegraphics[height=1.5em]{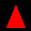}  

This is you facing right. \includegraphics[height=1.5em]{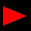}  

This is you facing down. \includegraphics[height=1.5em]{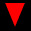}  

This is you facing left. \includegraphics[height=1.5em]{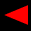}  

This is an empty tile. \includegraphics[height=1.5em]{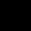}  

This is a key. \includegraphics[height=1.5em]{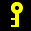}  

This is a locked door. \includegraphics[height=1.5em]{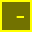}  

This is an unlocked door. \includegraphics[height=1.5em]{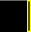}  

This is your goal. \includegraphics[height=1.5em]{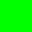}  

This is a wall. \includegraphics[height=1.5em]{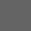}  
The possible actions are: 1. TURN LEFT, changing the direction to left 2. TURN RIGHT, changing the direction to right 3. MOVE FORWARD, moving forwards for one tile 4. PICK UP THE KEY, picking up the key, this is only possible when you are facing the tile with key on it 5. UNLOCK THE DOOR, this action is only possible when you have the key. Is action TURN LEFT the best action you can take? Please think step by step. Only give the answer in a new line in JSON format: \texttt{\{"reasoning": <REASONING>,"feedback": <FEEDBACK>\}}\\
where <FEEDBACK> is one of "YES" or "NO", <REASONING> is a string of your thinking steps. 

This is your image observation.

\includegraphics[width=0.4\textwidth]{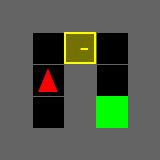}

\end{tcolorbox}

\clearpage

\begin{tcolorbox}[colback=red!5!white,colframe=red!75!black,title=Robomimic with Image Observation]
You are a robot engineer. You need to control a robot arm to lift an object. You need to observe the robot's current state and choose which action to do next in order to lift the object. You can move the robot in the X, Y, or Z direction, as well as open/close its gripper. You will also recieve a image as your observation. You can choose actions that move the gripper in the X, Y, Z direction, with additional control of the gripper. The format of the action is:\\n    [ resulting x, resulting y, resulting z, ignore, ignore, ignore, resulting gripper state]\\nIn the image, you will see you are facing the robot and a red cube (the object) is on the table. When dx < 0, the gripper moves away from you. When dx > 0, the gripper moves towards you. When dy < 0, the gripper moves towards the left of the picture. When dy > 0, the gripper moves towards the right of the picture. When dz < 0, the gripper moves towards the table. When dz > 0, the gripper moves away from the table.\\n Note on the gripper: Set grip to 1 to open the gripper and -1 to close the gripper. The robot's gripper should be closed if it is beginning to grasp the object, or when it is holding the object. When it is approaching the object, the gripper is open. If the robot gripper needs to be closed, you should continue to close the gripper, even if it is closed. Similarly, if the robot gripper needs to be open, you should continue to open the gripper, even if it is already open.

Your image observation is:

\includegraphics[width=0.4\textwidth]{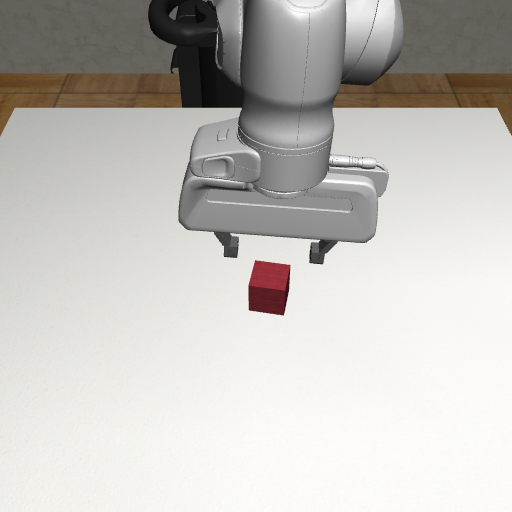}

Which action do you choose? Please think step by step. You should consider the position of the robot gripper and object, and how they are related to each other. For example, if the robot gripper is on the left of the object, you should consider moving the robot gripper to the right. Only give the answer in a new line in JSON format:

\texttt{\{"reasoning": <REASONING>, "action": <ACTION>\}}
where <ACTION> is the optimal action following the format of (dx, dy, dz), <REASONING> is a string of your thinking steps.
\end{tcolorbox}

\clearpage
\section{Domains and Dataset collection details}
\label{app:domain_dataset}

\paragraph{Multiple Optimal Actions and Expert Policies}
During the collection of our dataset, we need an expert policy.
However, in many domains, there might exist multiple optimal actions at the same time. 
In Cliff Walking, the expert policy is hand-coded and we enumerated all optimal actions for all states.
In MiniGrid Door key and MiniGrid Four Rooms, we use exhaustive search to enumerate all possible optimal actions.
In ALFWorld and HeirachyCraft, a built-in planner can serve as an expert policy.
Although these planners give an optimal action for each state, they do not yield the full set of all optimal actions.
Consequently, we cannot use these planners to compute the ground truth feedback for cases where there are multiple optimal actions.
For these two domains, we also use exhaustive search to enumerate all possible actions, and consider action optimal if after taking this action, there exists some sequence of remaining actions that reaches the goal in the same number of steps as the planner.

\paragraph{Preference and Binary Feedback}
We use the sampling policies only to sample states.
When constructing our preference and preference dataset for discrete domains, given a state, we enumerate all legal actions (for binary feedback) and all combinations of legal actions (for preference).

\paragraph{Robomimic}
In Robomimic, the continuous action space of the domain means we cannot enumerate all optimal actions. 
In Robomimic experiments, we use an imitation learning policy as an expert policy, which gives 100\% success rate in the task.
We also conducted an empirical study to decide the threshold distance for optimal actions. We injected different scales of random noise into the actions and inspect the task success rate and found 0.28 is the threshold of noise where bigger noise starts to negatively impact the success rate.
Therefore, we use this threshold as the biggest possible distance from the imitation learning policy action for an action to be optimal.
We also use this threshold for delta actions.

\clearpage
\section{Feedback Model Error Examples}
\label{app:response_examples}

In this section, we present some wrong responses from LLM.

\begin{tcolorbox}
[colback=red!5!white,colframe=red!75!black,title=Wrong response example in Cliff Walking Binary Feedback]
You are in position (0, 4). Is action DOWN the best action you can take?
\begin{verbatim}
{ "reasoning": "The agent's goal is to reach (3, 11) from (0, 4). 
Action DOWN moves to (1, 4). 
While this starts descending, reaching row 3 safely requires avoiding holes
in columns 1-10. Moving right from (0, 4) toward column 11 first ensures a
direct path to the goal without risking holes. Moving down now doesn't shorten
the path and delays aligning with the safe column (11). Thus, moving right
is better than down.",
"feedback": "NO"
}

\end{verbatim}
\end{tcolorbox}
In the example above, the LM assumes that they also need to keep away from the cliff, as well as give a wrong expectation of steps to the goal.

\begin{tcolorbox}[colback=red!5!white,colframe=red!75!black,title=Wrong response example in HiearchyCraft Action Advising]
\begin{verbatim}
<Prompt Omitted>
{"reasoning": "I need to find a way to get diamond, and I need to craft a furnace 
to smelt the ore. I have wood plank and stick, so I can craft a furnace. I also 
need to find iron ore or gold ore to smelt. I can search for wood at the swamp to 
get more wood and craft more items.", 
"action": "search for wood at swamp"}'
\end{verbatim}
\end{tcolorbox}

In the example above, the LM does not have knowledge of system dynamics. Even if it correctly reasons that it needs to collect more wood, it still picks the non-optimal action.
This action is not optimal because, moving to the forest and subsequently searching for wood leads to a shorter path to the goal.

\begin{tcolorbox}[colback=red!5!white,colframe=red!75!black,title=Wrong response example in Door Key Preference]
You need to solve Minigrid Doorkey.\\
Minigrid Doorkey involves crossing a gridworld, where you need to pick up the key, unlock the door and reach the goal. Do not hit the wall.\\
The state is described with an ASCII art. In the ASCII art, ``\#'' means wall. ``K'' means key. ``D'' means door. ``G'' means goal. ``.'' means empty tile. ``\^{ }'' means the agent who is facing upwards. ``V'' means the agent who is facing downwards. ``\textless{}'' means the agent who is facing leftwards. ``\textgreater{}'' means the agent who is facing rightwards.\\

The possible actions are:
\begin{enumerate}
  \item TURN LEFT, changing the direction to left, for example, the agent ``\^{ }'' will become ``\textless{}''.
  \item TURN RIGHT, changing the direction to right, for example, the agent ``\^{ }'' will become ``\textgreater{}''.
  \item MOVE FORWARD, moving forwards for one tile
  \item PICK UP THE KEY, picking up the key, this is only possible when you are facing the tile with key on it
  \item UNLOCK THE DOOR, this action is only possible when you have the key.

\end{enumerate}

You will be asked to compare two actions given a state and provide preference.\\

You are in the following state represented in ASCII art:
\begin{verbatim}
#####
#.#.#
#>D.#
#.#G#
#####
\end{verbatim}

The door is unlocked. You have the key.\\
Given \texttt{MOVE FORWARD} or \texttt{TURN LEFT}, which action is better? Please think step by step.\\
Only give the answer in a new line in JSON format:\\
\texttt{\{"reasoning": <REASONING>, "preference": <PREFERENCE>\}}\\
Where \texttt{<PREFERENCE>} is one of ``FIRST'' or ``SECOND'', \texttt{<REASONING>} is a string of your thinking steps.

\begin{verbatim}
{
  "reasoning": "In the current state, the agent is facing downwards and the door
  is unlocked. Moving forward is not a good option as it will lead to hitting the
  wall. Turning left is a better option as it allows the agent to change
  direction without moving and potentially see new possibilities to 
  reach the goal.",
  "preference": "SECOND"
}
\end{verbatim}
\end{tcolorbox}
In the wrong response example above, the LM misinterprets the observation and deems itself facing downwards, which leads to a wrong response in preferring the non-optimal action \texttt{TURN LEFT}.

\clearpage
\begin{tcolorbox}[colback=red!5!white,colframe=red!75!black,title=Wrong response example in Door Key Image Observation Action Advising]
<Prompt Omitted>

This is your image observation.

\includegraphics[width=0.4\textwidth]{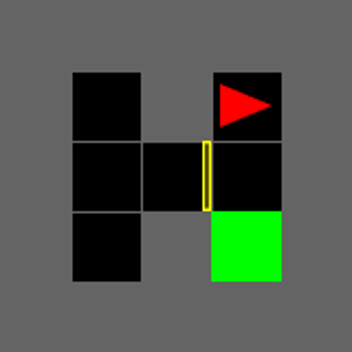}

\begin{verbatim}
{"reasoning": "I am currently facing right and need to move towards the 
goal which is directly ahead. Since there's no obstacle in my path, I 
can move forward.", "action": "MOVE FORWARD"}
\end{verbatim}

\end{tcolorbox}
In this wrong response example, the VLM misrecognises the position of the goal and disregards the wall ahead and decides to move forward, which will lead the agent to hit the wall.

\clearpage
\section{Reproducibility Statement}
The experiment details are converted throughout the paper as well as the appendix. For reproducibility, we included details in the dataset collection and evaluation pipelines in Section~\ref{sec:methodology}. We provide the prompt examples in Appendix~\ref{app:prompt_examples} and more implementation details in Appendix~\ref{app:domain_dataset}. 
We use models that are open-source and publicly available and we also provide a code and dataset appendix. We will also open-source our codebase.

\end{document}